\documentclass[journal,onecolumn]{IEEEtran}
\ifCLASSINFOpdf
\else
\fi
\usepackage{textcomp}

\usepackage{comment}

\usepackage{multirow}
\usepackage{soul}

\usepackage[noadjust]{cite}

\ifCLASSINFOpdf
  \usepackage{xcolor}
\else
  \usepackage{xcolor}
\fi
\usepackage[T1]{fontenc} 
\usepackage{amsmath}
\usepackage[cmintegrals]{newtxmath}
\usepackage{bm}

\usepackage{algorithm}
\usepackage{algpseudocode}

\usepackage{array}
\usepackage{url}
\usepackage[hidelinks]{hyperref}
\usepackage[utf8]{inputenc}
\usepackage[small]{caption}
\usepackage{graphicx}
\usepackage{amsmath}
\usepackage{booktabs}
\usepackage{subfigure} 

\ifCLASSINFOpdf
\else
\fi
\providecommand{\hypersetup}[1]{\relax}

\hypersetup{pdftitle={Bare Demo of IEEE\_lsens.cls for IEEE Sensors Letters},
pdfsubject={Typesetting},
pdfauthor={Michael D. Shell},
pdfkeywords={Class, IEEE, IEEE\_lsens, IEEE Sensors Letters, LaTeX, Typesetting, TeX}}

\usepackage{style}

\begin{document}
%
%
\title{Towards trustworthy multi-modal motion prediction: Holistic evaluation and interpretability of outputs}%

%
%
\author{\IEEEauthorblockN{Sandra~Carrasco~Limeros\IEEEauthorrefmark{1}\IEEEauthorrefmark{2}\IEEEauthorrefmark{3},
Sylwia~Majchrowska\IEEEauthorrefmark{2}\IEEEauthorrefmark{3},
Joakim~Johnander\IEEEauthorrefmark{3}\IEEEauthorrefmark{4}, 
Christoffer~Petersson\IEEEauthorrefmark{3}\IEEEauthorrefmark{5},
Miguel~Ángel~Sotelo\IEEEauthorrefmark{1}, and
David~Fernández~Llorca\IEEEauthorrefmark{1}\IEEEauthorrefmark{6}
}\\
\IEEEauthorblockA{\IEEEauthorrefmark{1}Computer Engineering Department, Polytechnic School, University
of Alcala, Madrid, Spain}\\
\IEEEauthorblockA{\IEEEauthorrefmark{2}AI Sweden, Göteborg, Sweden}\\
\IEEEauthorblockA{\IEEEauthorrefmark{3}Zenseact AB, Göteborg, Sweden}\\
\IEEEauthorblockA{\IEEEauthorrefmark{4}Department of Electrical Engineering, Linköping University, Linköping, Sweden}\\
\IEEEauthorblockA{\IEEEauthorrefmark{5}Chalmers University of Technology, Göteborg, Sweden} \\
\IEEEauthorblockA{\IEEEauthorrefmark{6}European Commission, Joint Research Centre, Seville, Spain}
\\
sandra.carrascol@uah.es, sylwia.majchrowska@ai.se, joakim.johnander@zenseact.com, christoffer.petersson@zenseact.com, miguel.sotelo@uah.es, david.fernandez-llorca@ec.europa.eu

\thanks{Authors' preprint, 2023.}
}
%
%

\markboth{CAAI Transactions on Intelligence Technology, 2023}%
{}

%

\IEEEtitleabstractindextext{%

}



\maketitle

\begin{abstract}
Predicting the motion of other road agents enables autonomous vehicles to perform safe and efficient path planning. This task is very complex, as the behaviour of road agents depends on many factors and the number of possible future trajectories can be considerable (multi-modal). Most prior approaches proposed to address multi-modal motion prediction are based on complex machine learning systems that have limited interpretability. Moreover, the metrics used in current benchmarks do not evaluate all aspects of the problem, such as the diversity and admissibility of the output.
In this work, we aim to advance towards the design of trustworthy motion prediction systems, based on some of the requirements for the design of Trustworthy Artificial Intelligence. We focus on evaluation criteria, robustness, and interpretability of outputs. First, we comprehensively analyse the evaluation metrics, identify the main gaps of current benchmarks, and propose a new holistic evaluation framework. We then introduce a method for the assessment of spatial and temporal robustness by simulating noise in the perception system. To enhance the interpretability of the outputs and generate more balanced results in the proposed evaluation framework, we propose an intent prediction layer that can be attached to multi-modal motion prediction models. The effectiveness of this approach is assessed through a survey that explores different elements in the visualization of the multi-modal trajectories and intentions. 
The proposed approach and findings make a significant contribution to the development of trustworthy motion prediction systems for autonomous vehicles, advancing the field towards greater safety and reliability.
\end{abstract}


%
\IEEEpeerreviewmaketitle

\section{Introduction}
The ability of human drivers to predict the motion of other road agents allows us to anticipate potentially dangerous situations and take preventive actions to minimise safety risks. It also allows humans to perform more efficient and comfortable maneuvers. It is therefore important that autonomous vehicles also have the capability to predict the motion of other road agents, so that they can apply predictive planning approaches and therefore behave in a more human-like manner. 

However, predicting future actions and motions of traffic participants is a very complex task, as the behaviour of road agents is influenced by many different variables and interactions \cite{Rudenko2020}, \cite{Biparva2022}. Furthermore, despite the fact that traffic environments are well structured (e.g. street layout, traffic rules), the number of possible future trajectories for each past trajectory for each agent can be considerable, whether for pedestrians, cyclists or vehicles. That is, the problem is multi-modal in nature. 

In order to handle this complexity, most of the computational approaches proposed to address multi-modal motion prediction rely on very complex machine learning models which are far from being interpretable. These models are not at human scale and suffer from the characteristic of opacity (i.e., black-box models). Besides, there is no consensus on the most important metrics that should be used to evaluate their performance. Different benchmarks propose different metrics and, in most cases, focus mainly on accuracy, omitting some other relevant aspects such as robustness, diversity, or compliance with traffic rules.  

Furthermore, in recent years it has been increasingly accepted that the design of complex learning-based systems must follow certain rules to ensure compliance, not only with traditional safety requirements, but also with general ethical grounds. This approach has recently been referred to as \emph{Trustworthy AI}.  It is a concept that encompasses multiple ethical principles, requirements, and criteria to guarantee that AI systems are designed following a human-centered approach and committed to social good \cite{Llorca2022}. The development of human-centric AI is now a common trend worldwide. For example, at EU level, the High Level Expert Group on AI (AI HLEG) appointed by the European Commission (EC) defined the main horizontal requirements \cite{HLEGAI} and criteria \cite{ALTAI} to develop trustworthy AI systems, including elements such as human oversight, robustness and safety, privacy, and data governance, transparency, fairness, well-being, and accountability. In April 2021, the EC presented the \emph{Proposal for a regulation lying down harmonized rules on AI} (the AI Act \cite{AIAct}) which imposes a set of requirements for AI systems used in high-risk scenarios. Among other things, the AI Act states that the relevant accuracy metrics shall be fit for purpose, and that technical measures shall be put in place to facilitate the interpretation of the outputs of AI systems. At the US level, albeit with a different focus (“algorithms” and “automatic decision systems” instead of "AI systems”, and “critical decisions” instead of “high-risk scenarios”), the \emph{Algorithmic Accountability Act} \cite{US-AAA2022} was introduced in the US Senate and the House of Representatives in February 2022, which also imposes specific requirements on impact assessment, documentation and performance evaluation of automated critical decision systems. 

\begin{figure}[htb!]
    \centering
    \includegraphics[width=\linewidth]{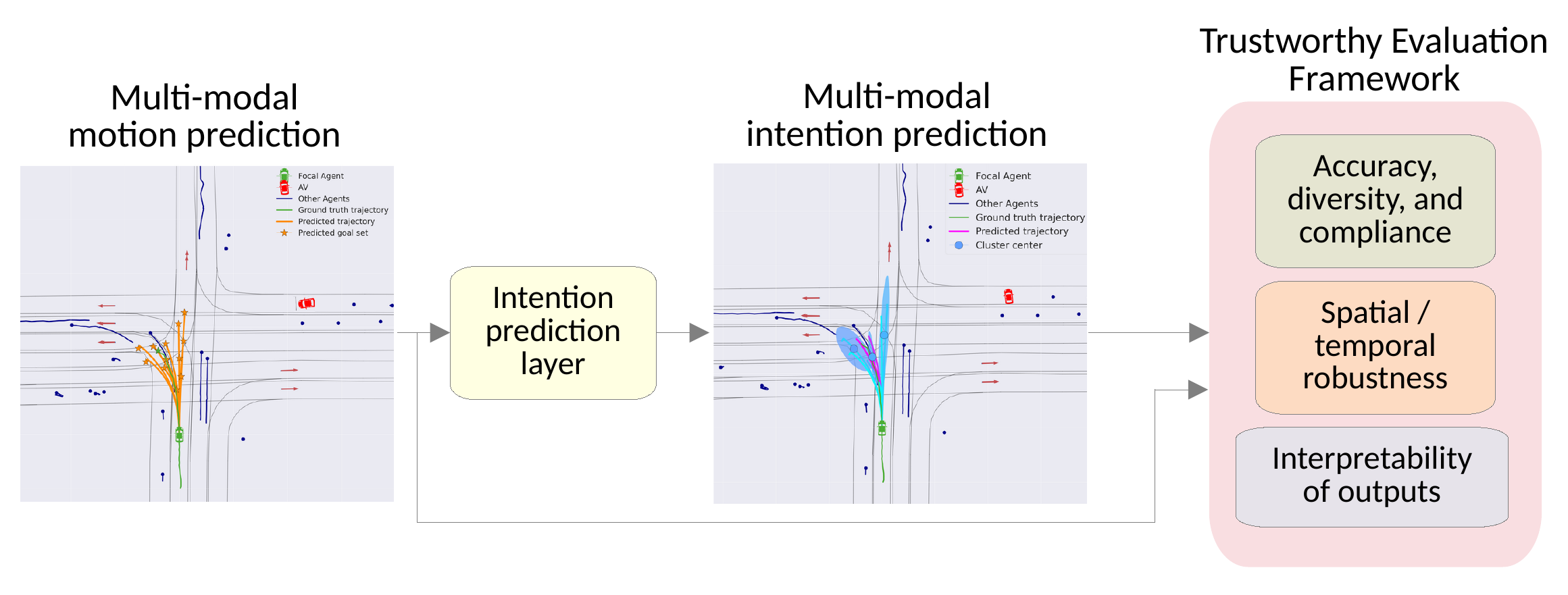}
    \caption{Towards trustworthy multi-modal motion prediction. The intention prediction layer enhances the interpretability of the results. The proposed evaluation framework includes a holistic set of metrics and a robustness analysis.}
    \label{fig:abstract}
\end{figure}

In this work, we aim to advance towards the design of trustworthy multi-modal motion prediction systems, based on some of the aforementioned requirements. More specifically, we focus on evaluation criteria, including robustness, and interpretability of outputs. The main contributions can be summarized as follows (see Figure \ref{fig:abstract}): 

\begin{itemize}
    \item We comprehensively analyse the evaluation metrics, identifying the main gaps and proposing a new holistic evaluation framework which jointly takes into account accuracy, diversity and compliance with traffic rules (admissibility). 
    \item We formulate a new method to assess the spatial and temporal robustness of multi-modal motion prediction by simulating noise in the perception system. 
    \item We propose a new intent prediction layer that can be attached to multi-modal trajectory prediction models to enhance the interpretability of the outputs and generate more balanced results in the proposed holistic evaluation framework.
    \item We assess the interpretability of the outputs and the evaluation framework by means of a survey that explores different elements in the visualization of the multi-modal trajectories and intentions. 
\end{itemize}

The proposed study is carried out on two of the most widely used datasets in the field (Argoverse \cite{Chang_2019_CVPR} and nuScenes \cite{nuscenes2019}), and using two state-of-the-art multi-modal prediction approaches (DenseTNT \cite{Gu2021DenseTNTET} and PGP \cite{deo2021multimodal}).

The remaining of this paper is organized as follows. Section~\ref{sec:related_work} presents a literature review on multi-modal motion prediction. 
In Section~\ref{sec:methods}, we analyse the current evaluation methods for multi-modal motion prediction models and propose a more comprehensive evaluation framework. 
Section \ref{sec:robustness} presents the assessment of the robustness of two state-of-the-art multi-modal prediction models in nuScenes and Argoverse.  
In Section~\ref{sec:intention} we describe the intention prediction layer to cluster the output predictions into high-level behaviours to enhance the interpretability of the output. 
Section~\ref{sec:survey} provides the study on how multi-modality impacts interpretability by means of a survey. 
Finally, conclusions and next steps are presented in Section~\ref{sec:discussion}.

\section{Multi-modal motion prediction}
\label{sec:background}

\subsection{Methods}
\label{sec:related_work}

Recently, many different approaches were proposed to deal with multi-modal motion prediction~\cite{tang2019mfp,Trajectronpp,konev2021motioncnn,liu2021multimodal,MultiPathpp,gilles2022thomas}. The state-of-the-art methods rely on graph models~\cite{ vectornet2020,tnt,Gu2021DenseTNTET,bib:Carrasco2021,deo2021multimodal}, since traffic scenarios can be naturally represented as graphs. The graph structure is flexible and efficient because it allows direct and explicit representation of interactions through edges. Usually graph-based models express the features of agents (e.g., vehicles, pedestrians, bicycles) in the nodes, and the relationships between agents as edges~\cite{bib:Carrasco2021}. 
There are several architectural design decisions that must be made in order to effectively represent the input data, model the interaction and finally represent the output trajectory distributions.

\subsubsection{Representation of high-definition maps}
Methods for motion prediction need to effectively represent both geometric information (static scene elements) and traffic agents (dynamic scene elements). The standard raster representation encodes the world as a stack of bird's eye view (BEV) images, also called high-definition (HD) maps~\cite{mtp,tang2019mfp,Trajectronpp,spagnn}. This approach is straightforward as all the different types of input information (e.g., road configuration, state history of agents, spatial relationships) are unified in a multi-channel image, allowing the use of a Convolutional Neural Network (CNN). However, the this approach is limited by narrow receptive field of standard CNNs. Hence, rasterized representations has difficulties in modeling long-distance interactions.

Alternatively, less expensive graph-based representation can be used in the form of polylines (e.g., lanes, crosswalks, boundaries), which represent piecewise linear segments~\cite{MultiPathpp,vectornet2020,tnt,Gu2021DenseTNTET,Wang_2022_CVPR,Khandelwal2020WhatIfMP}. In this representation, long-range dependencies are effectively and efficiently modelled. For example, VectorNet~\cite{vectornet2020,tnt,Gu2021DenseTNTET} represents the lane segments and agents histories as vectors connected end-to-end. As an improvement, Lane-based Trajectory Prediction (LTP) method~\cite{Wang_2022_CVPR} encodes the map through connected sliced lane segments to more precisely model the intention. Other approaches~\cite{GOHOME,Lanegcn,LaneRCNN,deo2021multimodal} construct an additional lane graph, leveraging temporal and spatial correspondence of the vectorized scene context.

\subsubsection{Interaction modelling}
There are two types of interactions that need to be modelled. First, the encoding of temporal sequential data, which is typically accomplished via Recurrent Neural Networks (RNNs) -- such as the Gate Recurrent Unit (GRU)~\cite{gru} or the Long Short-Term Memory (LSTM)~\cite{lstm} -- or temporal convolutions. Second, the  interactions between the relevant agents and the environment is modelled via an attention mechanism~\cite{NIPS2017_3f5ee243}. Interaction modelling is closely related to the method used for scene encoding.

Based on the polyline representation, VectorNet~\cite{vectornet2020,tnt,Gu2021DenseTNTET} utilises only self-attention modules to directly learn the interactions between all the sub-graphs in the environment. Further, goal-oriented lane attention~\cite{Luo2020ProbabilisticMT} emphasizes
the relationship between agents and lanes. The split-joint attention mechanism, used by LaneGCN~\cite{Lanegcn}, captures the complex topology of lane graphs and long range dependencies. 
Deo et al.~\cite{deo2021multimodal} propose Prediction via Graph-based Policy (PGP) model, where interactions are modelled via lane-graph traversals. This approach combines discrete policy roll-outs with a lane-graph subset decoder, conditioning each prediction on the driver's goals. Finally, the network predicts trajectories by selectively attending to node encodings along paths traversed by the policy and a sampled latent variable.

\subsubsection{Multi-modal output}
To account for environmental uncertainty, models predicting future trajectories should represent multi-modal paths~\cite{mtp,Luo2020ProbabilisticMT,PhanMinh2020}. 
This can be represented by implicitly modelling multi-modality as latent variables or explicitly proposing to generate multiple trajectory proposals. 
The first approach uses Gaussian Mixture Models (GMMs)~\cite{Chai2019MultiPathMP} or Mixture Density Networks (MDNs)~\cite{Makansi_2019_CVPR} to generate distributions for possible trajectories. In next step, it uses Variational Autoencoders (VAEs), recently also their modified version as Conditional Variational Autoencoder (CVAEs)~\cite{Cao2021SpectralTG}, or
GANs~\cite{Gupta2018SocialGS,Chai2019MultiPathMP,Lee2017DESIREDF} to sample various future modes from latent variables. 
The main drawback of this approach is that the obtained predictions cannot be interpreted unambiguously, which lowers the understanding of the model's predictions.

On the other hand, the proposal-based methods design various possible proposals and thus separately solve the tasks of intention prediction and motion prediction. Some methods~\cite{tnt,Gu2021DenseTNTET} sample proposal points around the lane centerline to capture detailed information. The proposed sampling is based on carefully developed rules, the restriction on sharing proposals between different agents is one of them. In contrast,~\cite{Wang_2022_CVPR} uses lane segments from the map as proposals that can clearly describe the fine-grained intentions of agents and be shared globally.

From predicted paths, the most reliable trajectories are selected. Here, a variation of Non-maximum Suppression (NMS) algorithm is widely used. The traditional version of this method is employed in TNT~\cite{tnt}, in which each trajectory is sorted according to the method scoring process. Then, the distance between different trajectories is calculated and a diverse set of trajectories with high scores is selected. Similar to TNT, LaneRCNN~\cite{LaneRCNN } treats a lane segment as an anchor and output each anchor’s probability, using NMS to remove too close duplicate goals. The solution has its drawbacks, as a fixed threshold does not allow the model to maintain the balance between accuracy and multi-modality of output. In contrast, DenseTNT~\cite{Gu2021DenseTNTET} is an anchor-free goal-based model, which generates a set of predicted goals without relying on the heuristic anchors.

\subsection{Datasets}
\label{sec:datasets}
Experiments presented in subsequent sections are conducted on two publicly available trajectory forecasting benchmarks: Argoverse v1.1 Motion Forecasting Dataset~\cite{Chang_2019_CVPR} and nuScenes~\cite{nuscenes2019}. These two datasets are the most widely used in the motion prediction task. The datasets differ from each other in many parameters, e.g., number of scenes, total driving time, number of objects, interaction complexity, or prediction time horizon.
The prediction time horizon, in particular, plays a crucial role. Long-term forecasting is inherently more difficult, and multi-modal forecasting appears to be more successful in this case. Correctly determining the driver's intentions and the future trajectory of the vehicle is essential for safe planning. It strongly depends on the overall quality of the dataset on which the algorithms were trained, so it is important to select appropriate data for future applications..

\textbf{Argoverse}~\cite{Chang_2019_CVPR}, released in 2019 by Argo AI, is composed of 323~557 real-world driving sequences. Driving scenarios were collected in two American cities: Miami and Pittsburgh. Each example consists of 2 seconds of historical state and 3 seconds of future state, which is sampled with a frequency of 10 Hz. The entire collection of scenarios totals 320h. In each scene, one actor of interest is specified whose future movement is to be predicted. For training and validation, focal agent history, location histories of nearby (social) actors, together  with HD map features are also provided. The semantic HD map provided is simplistic and consists of lane-based polylines.

\textbf{nuScenes}~\cite{nuscenes2019}, released in 2020 by Motional, is a large-scale data collection for  multi-agent trajectory forecasting with 1,000 scenes recorded in Boston and Singapore, where right-hand and left-hand traffic rules apply respectively.
Each scene is annotated at a frequency of 2 Hz and is 20s long. It contains up to 23 semantic object classes, including vehicles, bicycles and pedestrians as possible tracked agents, as well as HD semantic maps with 11 annotated layers.
For this dataset, each agent has 2 seconds of observed trajectory and the prediction horizon is set to 6 seconds.

Overall,  Argoverse---which tends to represent short, independent episodes---is more than fifty times larger in terms of total time than nuScenes.  Despite the impressive size of the Argoverse dataset, previous works have observed that the recorded trajectories mostly represent straight-line trajectories within a full 5-second window~\cite{Khandelwal2020WhatIfMP,Sanmin22}. This makes the dataset much less challenging and diverse than nuScenes. The datasets differ also in the schematic representation of the HD maps. Argoverse presents a much simpler representation, while nuScenes is more comprehensive but complex. We simplified  nuScenes representations to focus on the most representative information for the conducted study.

\section{Evaluation framework}
\label{sec:methods}


\label{sec:metrics}
In this section, we revise the evaluation procedures of multi-modal prediction approaches.
The presented metrics are differentiated between measures of precision, diversity, and admissibility (see Table \ref{tab:evaluation}). 
Diversity refers to the degree of coverage of the output distribution. We seek to have diversity across distribution modes rather than having multiple candidates representing a single intention (mode). It is important to evaluate, however, the admissibility of such predictions. They need to comply with the traffic rules.  

\textbf{Accuracy evaluation}:
\begin{itemize}
    \item Best-of-K Average Displacement Error (minADE): The minimum point-wise L2 distance to the ground-truth trajectory over all predicted trajectories. 
    \item Scene minADE: Used when considering the prediction of multiple agents simultaneously. It measures the joint minimum L2 distance between predictions and ground-truth over all the agents in the scene. 
    \item Best-of-K Final Displacement Error (minFDE): Lowest L2 distance between the K predicted endpoints and the ground truth endpoint at the prediction horizon over all predicted trajectories.
    \item Scene minFDE: To some extent, similar to scene minADE.  It measures the joint minimum FDE over the whole scene.
    \item Best-of-K Miss Rate (MR): Percentage of predictions whose maximum pointwise L2 distance between the prediction and ground truth is greater than a threshold. For Argoverse and nuScenes benchmarks a threshold of 2.0m is considered. However, they define this metric in different ways. Whereas the nuScenes benchmark uses the above definition, Argoverse computes MR as the number of scenarios where none of the forecasted trajectories are within 2m of the  ground-truth endpoint.
    Waymo~\cite{Ettinger_2021_ICCV} differentiates between lateral and longitudinal thresholds, which scale depends on future time horizon and initial speed.
    \item Heading error: The difference in the heading angle between the predicted trajectory and the ground truth trajectory at the final point.
    \item Mean Average Precision (mAP)~\cite{Ettinger_2021_ICCV}: The area under the precision-recall curve by applying confidence score thresholds across a validation set. They use the same definition of a miss as defined for MR, considering any missed prediction as false positive. Only one true positive is allowed for each prediction -- assigned to the highest confidence prediction.  
   They report the final mAP averaged over eight different semantic buckets or driving behaviors, i.e., straight, straight-left, straight-right, left, right, left u-turn, right u-turn, and stationary. This makes the evaluation more balanced, since some of these trajectory shapes are much more infrequent than others. 
    \item Soft Mean Average Precision (Soft mAP)~\cite{Ettinger_2021_ICCV}: Similar to mAP, but additional matching predictions, other than the highest confidence one, are not penalized. 
    
\end{itemize}
Multi-modal prediction models are usually benchmarked using Best-of-K metrics. These metrics, although useful for deterministic regressors, only take into account a single output of an arbitrary number K, representing a limited part of the model output distribution. Hence, they are not able to compare the distributions produced by multi-modal models, neglecting the assessment of variance and multi-modality. 

Furthermore, these metrics only assess the quality of the underlying marginal distribution per agent. Best-of-K ADE takes the trajectory sample that is closest to the ground truth of each agent in an independent manner. In this way, we are not measuring scene consistency in the predictions. It would be possible to have a low minADE by predicting high-entropy distributions that are not consistent at the scene level.  
In  \cite{casas2020implicit}, the authors propose scene-level sample metrics to assess how well the output modes capture the joint distribution over future motions.  

\textbf{Probabilistic evaluation:} In order to make a fair assessment of the probabilistic capabilities of a model, we need to measure its ability to capture the underlying uncertainty distribution of future motions. In the following, we list some of the metrics used in the literature \cite{Chang_2019_CVPR} to evaluate probabilistic prediction models. Being \textit{p} the probability of the best forecasted trajectory:   
\begin{itemize}
    \item Probabilistic minimum Final Displacement Error  (p-minFDE): Similar to minFDE. It adds $min(-\log p, -\log 0.05)$ to the endpoint L2 distance.
    \item Probabilistic minimum Average Displacement Error  (p-minADE): Similar to minADE. It adds $min(-\log p, -\log 0.05)$ to the average L2 distance.
    \item Probabilistic Miss Rate (p-MR): Similar to Miss Rate. It takes $1.0-p$ as the contribution, instead of 0.0, when the L2 distance between the maximum pointwise L2 distance between prediction and ground truth is greater than the threshold. 
    \item Brier minimum Final Displacement Error  (brier-minFDE) \cite{Brieer}:   This metric is similar to minFDE, but we add $(1.0-p)^2$ to the endpoint L2 distance.  
    \item Brier minimum Average Displacement Error  (brier-minADE) \cite{Brieer}:  This metric is similar to minADE, but we add $(1.0-p)^2$ to the average L2 distance.  
    \item Negative Log Likelihood (NLL) \cite{Ivanovic2019}: The average negative log likelihood of the ground truth trajectory as determined by a kernel density estimate over output samples at the same prediction timestep.
\end{itemize}

It is important to note that different benchmarks consider a different number of modes to evaluate prediction accuracy. For instance, Argoverse and Waymo consider the 6 most likely trajectories, whereas nuScenes takes 5 and 10 modes.
For probability-based metrics in Argoverse, they take the 6 most likely trajectories and normalize the probabilities before computing the metrics.

There is also a difference in the prediction horizon considered. Both Argoverse and nuScenes provide 2s of trajectory history. However, Argoverse prediction time horizon is 3s, whereas nuScenes considers 6s.  Waymo \cite{Ettinger_2021_ICCV} provides tracks for the past 1 second and considers three different evaluation times: 3, 5, and 8 seconds into the future.

In order to sort their leaderboards, there is no consensus on the most important metric. nuScenes ranks the different approaches according to minADE$_5$ (K=5). Argoverse sorts them according to brier-minFDE at K=6, assuming a uniform distribution for approaches that do not provide a probability associated to the predictions. In previous competitions, they used MR and minFDE at K=6. Waymo leaderboard ranking is built based on the average Soft mAP across evaluation times, while MR is used as a secondary metric.


\textbf{Diversity evaluation.} Evaluating the diversity of multiple predictions is required to properly assess the multi-modal predictive capabilities. However, this evaluation is far less explored in the literature, resulting in models that, although obtaining good results in terms of accuracy, suffer from mode collapse and do not show real multi-modality. 
In the following, we list metrics used to evaluate the diversity of the predictions:
\begin{itemize}
    \item Lateral diversity metrics: The authors of \cite{deo2021multimodal} report the average number of different lanes reached as a measure of lateral diversity, as well as the variance of the final heading for the different output modes. \cite{Sanmin22} defines \textit{minLaneFDE}, which captures both the quantity and quality of diversity of multiple outputs based on the centerlines of reference lanes. It computes the minimum value among the L2 distance between centerlines of possible lane candidates and each predicted mode. 
    \item Longitudinal diversity metrics: Deo \textit{et.al}~\cite{deo2021multimodal} reports the variance of the average speeds and accelerations for the different output modes.
    \item Ratio of avgFDE to minFDE (RF): Proposed in \cite{diverseandadmissible}, it measures the spread of the predictions in Euclidean distance.
\end{itemize}

A large average L2 error implies that the predictions are spread out, while a small minimum L2 error implies that at least one of the predictions has high precision. The authors in~\cite{diverseandadmissible} follow this intuition and propose to compute the ratio of avgFDE to minFDE to capture diversity in the output. However, this metric can fail to differ between longitudinal and lateral diversity in some scenarios. In addition, this metric may be better for models that are worse in terms of accuracy and admissibility when most modes are failing modes. A straightforward way to assess lateral diversity can be to measure the variance of the final heading for the K possible outputs. When having access to the lane information, the average number of final lanes reached is a good measure of mode diversity. On the other hand, longitudinal diversity can be measured by the variance of the final speed and acceleration for the K outputs.

\textbf{Admissibility evaluation.}  Finally, assessing the compliance of the predictions with the driving scene is essential to evaluate the quality of the output and to ensure a safe motion planning in complex driving scenarios. In the following, we list the metrics used to evaluate the admissibility of the predictions: 
\begin{itemize} 
    \item Off-road rate:  Computes the fraction of trajectories that are off-road, outside the drivable area. 
    \item Drivable Area Occupancy (DAO) \cite{diverseandadmissible}: Measures the proportion of pixels occupied by the predicted trajectories within the drivable area. This metric can be used together with RF to get a more compliant measure of diversity.
    \item Drivable Area Compliance (DAC) \cite{Chang_2019_CVPR}: Measures extreme off-road predictions that are not admissible.  If a model produces K modes for future trajectories and M of those going off the drivable area at any point, the DAC for that model would be (N - M)/N.
    \item Scene consistency rate (SCR)~\cite{casas2020implicit}: Measures the percentage of predicted samples that collide (overlap) in the scene  in order to evaluate social-consistency. A collision is detected by comparing the IOU between the future BEV-defined bounding boxes of each pair of agents in the scene with a small IOU threshold.
    \item Overlap Rate~\cite{Ettinger_2021_ICCV}: Similar to the previous metric, it takes the highest confidence prediction from each agent and computes the total number of overlaps divided by the total number of agents. A single overlap is defined if any of these trajectories overlaps at any time with any other agent at the prediction time step. 
\end{itemize} 
Off-road rate, DAO, and DAC are useful for measuring the proportion of modes that go off-road. The last two metrics evaluate the ability of the model to capture social interactions. However, to the best of our knowledge, there are no metrics that assess compliance with traffic rules. To this end, we propose to measure the ratio of modes going in the oncoming traffic direction $OTD = \frac{count(traj_{OTD})}{K}$.

Autonomous driving is a high-stake and safety-critical application. As such, it is of utmost importance to  thoroughly evaluate each of its intermediate systems, including the trajectory prediction stage that will be an essential input for safe and efficient planning. A comprehensive evaluation and interpretation of the performance of the prediction model is needed in terms not only of precision, but also diversity and admissibility. 
 Simultaneous examination of all these dimensions provides a holistic evaluation framework for the assessment of multi-modal motion prediction.
In practice, the choice of which metrics to evaluate will depend on the particular needs of the system and the specific application context. Our framework serves as a guide to help the user make an informed decision about which metrics are most relevant to their needs and to provide a basis for comparison with other models.
 
 \begin{table*}[]
\centering
\caption{Reviewed metrics categorized in terms of Accuracy, Diversity, and Admissibility. \textit{p} corresponds to the probability of the best forecasted trajectory. Symbol \textit{-} refers to \textit{no units}, \textit{\#} refers to \textit{number of}}.
\label{tab:evaluation}
\resizebox{\textwidth}{!}{%
\begin{tabular*}{\textwidth}{|p{15pt}<{\raggedright}|p{55pt}<{\raggedright}|p{80pt}<{\raggedright}|p{20pt}<{\raggedright}|p{285pt}<{\raggedright}|}
\toprule
                               &                                         & \textbf{Metric}                                                                                 & \textbf{Unit }           & \textbf{Description}                                                                                                                                                                                                                                                                                                                     \\ \midrule
\multirow{14}{*}{\rotatebox[origin=c]{90}{\textbf{Accuracy}}}     & \multirow{6}{*}{Non-probabilistic }     & Minimum Average Displacement Error  (minADE)& m ($\downarrow$) & The average of pointwise L2 distances between the predicted trajectory and ground truth over the k  most likely predictions.                                                                                                                                                                                                    \\ \cmidrule(l){3-5} 
                               &                                         & Minimum Final Displacement Error  (minFDE)   & m ($\downarrow$) & The L2 distance between the final points of the prediction and ground truth,  over the k most likely predictions.                                                                                                                                                       \\ \cmidrule(l){3-5} 
                               &                                         & Miss Rate (MR)                             & \% ($\downarrow$)                           & Percentage of scenarios where none of the forecasted trajectories for an agent are within 2.0 meters of ground truth according to endpoint error.                                                                                                                                                                               \\ \cmidrule(l){3-5} 
                               &                                         & Heading error                                                                          & rad  ($\downarrow$)                         & Difference in the heading angle between the predicted trajectory and the ground truth trajectory at the final point.                                                                                                                                                                                                            \\ \cmidrule(l){3-5} 
                               &                                         & Mean Average Precision (mAP)               & -  ($\uparrow$)                           & Area under the precision-recall curve by applying confidence score thresholds across a validation set. True and false positives are defined using the definition of MR.  Only one true positive is allowed for each agent, assigned to the highest confidence prediction. \\ \cmidrule(l){3-5} 
                               &                                         & Soft Mean Average Precision (Soft mAP)     & -   ($\uparrow$)                          & Similar to mAP, where additional matching predictions, other than the highest confidence one, are not penalized.                                                                                                                                                                                                                \\ \cmidrule(l){2-5} 
                               & \multirow{2}{*}{Scene-level}            & Scene minADE                                                                           & m ($\downarrow$) & Joint minimum L2 distance between predictions and ground-truth over all the agents in the scene.                                                                                                                                                                                                                                \\ \cmidrule(l){3-5} 
                               &                                         & Scene minFDE                                                                           & m ($\downarrow$) & Joint minimum L2 distance between the K predicted endpoints and the ground truth endpoint at the prediction horizon over all predicted trajectories.                                                                                                                                                                            \\ \cmidrule(l){2-5} 
                               & \multirow{6}{*}{Probabilistic }         & p-minADE                                                                               & m ($\downarrow$) & Similar to minADE, adding $ min(-log(p), -log(0.05)) $ to the average L2 distance.                                                                                                                                                                                                                                                  \\ \cmidrule(l){3-5} 
                               &                                         & p-minFDE                                                                               & m ($\downarrow$) & Similar to minFDE, adding $min(-log(p), -log(0.05)) $to the endpoint L2 distance.                                                                                                                                                                                                                                                 \\ \cmidrule(l){3-5} 
                               &                                         & p-MR                                                                                   & \%  ($\downarrow$)                           & Similar to Miss Rate, taking $(1.0 - p)$ as the contribution, instead of 0.0, when the endpoint error for the best forecasted trajectory is less than 2.0m.                                                                                                                                                                       \\ \cmidrule(l){3-5} 
                               &                                         & brier-minADE                                                                           & m ($\downarrow$) & Similar to minFDE, adding $(1.0 -p)^2$ to the endpoint L2 distance.                                                                                                                                                                                                                                                               \\ \cmidrule(l){3-5} 
                               &                                         & brier minFDE                                                                           & m ($\downarrow$) & Similar to minADE, adding $(1.0 -p)^2$  to the average L2 distance.                                                                                                                                                                                                                                                               \\ \cmidrule(l){3-5} 
                               &                                         & Negative Log Likelihood (NLL)               &  - ($\downarrow$)                              & The average negative log likelihood of the ground truth trajectory as determined by a kernel density estimate  over output samples at the same prediction timestep.                                                                                                                                                           \\ \midrule
\multirow{6}{*}{\rotatebox[origin=c]{90}{\textbf{Diversity}}}     &                                         & Ratio of avgFDE to minFDE (RF)           & m/m   ($\uparrow$)                        & Ratio avgFDE to minFDE as a measure of the spread of the predictions in Euclidean distance.                                                                                                                                                                                                                                     \\ \cmidrule(l){2-5} 
                               & \multirow{3}{*}{Lateral}      & \# Lanes reached                                                                       & \#    ($\uparrow$)                        & Average number of different lanes reached.                                                                                                                                                                                                                                                                                      \\ \cmidrule(l){3-5} 
                               &                                         & $\sigma_\mathrm{yaw}^2$                                      & rad$^2$   ($\uparrow$)                       & Variance of the final heading for the different output modes.                                                                                                                                                                                                                                                                   \\ \cmidrule(l){3-5} 
                               &                                         & LaneFDE                                                                                & m ($\downarrow$) & Minimum value among the L2 distance between centerlines of possible lane candidates and each predicted mode.                                                                                                                                                                                                                    \\ \cmidrule(l){2-5} 
                               & \multirow{2}{*}{Longitudinal} & $\sigma_\mathrm{v}^2$                                        & m/s$^2$  ($\uparrow$)                        & Variance of the average speeds.                                                                                                                                                                                                                                                                                                 \\ \cmidrule(l){3-5} 
                               &                                         & $\sigma_\mathrm{a}^2$         &                                m/s$^2$ ($\uparrow$)                         & Variance of the average accelerations.                                                                                                                                                                                                                                                                                          \\ \midrule
\multirow{6}{*}{\rotatebox[origin=c]{90}{\textbf{Admissibility}}} & \multirow{3}{*}{Drivable Area}     & Off-road rate                                                                          & \%  ($\downarrow$)                           & Fraction of trajectories that are outside the drivable area.                                                                                                                                                                                                                                                                    \\ \cmidrule(l){3-5} 
                               &                                         & Drivable Area Occupancy (DAO)                                                           & \#   ($\uparrow$)                         & Proportion of pixels occupied by the predicted trajectories within the drivable area.                                                                                                                                                                                                                                           \\ \cmidrule(l){3-5} 
                               &                                         & Drivable Area Compliance  (DAC)                                                            & \#   ($\uparrow$)                         & If a model produces k possible future trajectories and m of that exit the drivable area at some point, the DAC for that model would be (n-m)/n.                                                                                                                                                                                 \\ \cmidrule(l){2-5} 
                               & \multirow{2}{*}{Social interactions}    &  Scene consistency rate  (SCR)                                                              & \%  ($\downarrow$)                           & Percentage of predicted samples that overlap in the scene looking at the IOU between the future BEV-defined bounding boxes of each pair of agents.                                                                                                                                                                              \\ \cmidrule(l){3-5} 
                               &                                         & Overlap Rate                                                                           & \%  ($\downarrow$)                           & Number of overlaps divided by the total number of agents. A single overlap is defined if any of these trajectories overlaps at any time with any other agent at the prediction time step.                                                                                                                                       \\ \cmidrule(l){2-5} 
                             & \multirow{1}{*}{Traffic rules }                        &  Oncoming Traffic Direction (OTD)                                                      & \%   ($\downarrow$)                          & Ratio of modes going in the oncoming traffic direction.                                                                                                                                                                                                                                                                         \\ \bottomrule
\end{tabular*}%
}
\end{table*}
 


\section{Robustness Analysis}
\label{sec:robustness}

\begin{table*}[ht!]
\centering
\caption{Robustness analysis of PGP in NuScenes dataset. K=10 for all experiments.}
\begin{tabular}{|cc|ccccc|}
\hline
    Object   & Experiment     & minADE$_5$    & minADE$_{10}$    & miss rate$_5$    & miss rate$_{10}$   & BC \\
\hline
  &    \begin{tabular}[c]{@{}c@{}}Baseline \\ Recall 100\%, K=10\end{tabular}          & \textbf{1.30}   &\textbf{ 0.97}      & \textbf{0.52}      &\textbf{0.34 }    & \textbf{1.91} \\
  \hline
\multirow{3}{*}{Lanes}  &Recall 95\%      & 1.36   & 1.01      & 0.55      & 0.38    & 2.51 \\
                        &Recall 90\%      & 1.41     & 1.04    & 0.57      & 0.40     & 3.0 \\
                        &Recall 80\%      & 1.54     & 1.12    & 0.60      & 0.43     & 4.01 \\
\hline
\multirow{5}{*}{Dynamic agents} &Recall 95\%    & 1.30   & 0.97    & 0.52      &0.34       & 1.93 \\
                                &Recall 90\%    & 1.31   & 0.97    & 0.53      &0.35       & 1.93 \\
                                &Recall 80\%    & 1.32   & 0.97    & 0.53      &0.34       & 1.94 \\
                &Recall 50\% temporal   & 1.34   & 0.98    & 0.54      &0.36       & 1.96 \\
                        &No detected agents     & 1.42   & 1.01    & 0.58      &0.39       & 2.07\\
\hline
\end{tabular}
\label{tab:robustness}
\end{table*}

\begin{table*}[]
\centering
\caption{Robustness analysis of DenseTNT in Argoverse dataset. K=12 for all experiments.}
\label{tab:robustnessDenseTNT}
 
\begin{tabular}{|cc|cccccc|}
\hline
Object                           & Experiment                                                     & minADE  & minFDE  & Miss Rate  & AvgFDE  & RF  & DAC  \\ 
\hline
                                         & \begin{tabular}[c]{@{}c@{}}Baseline \\ Recall 100\%, K=12\end{tabular} & \textbf{0.763 }          & \textbf{1.078 }          & \textbf{0.043  }            & \textbf{5.447}           & 5.053       & \textbf{0.971 }       \\ 
                                         \hline
\multirow{4}{*}{Lanes }          & Recall 95\%                                                            & 0.777           & 1.110           & 0.049              & 5.499           & 4.954       & 0.971        \\
                                         & Recall 90\%                                                            & 0.799           & 1.151           & 0.059              & 5.577           & 4.849       & 0.971        \\
                                         & Recall 80\%                                                            & 0.858           & 1.281           & 0.084              & 5.823           & 4.546       & 0.969        \\
                                         & Recall 70\%                                                             & 0.948           & 1.478           & 0.120              & 6.144           & 4.156       & 0.967        \\
                                         \hline
\multirow{6}{*}{Dynamic agents} & Recall 95\%                                                            & 0.763           & 1.081           & 0.043              & 5.459           & 5.049       & 0.971        \\
                                         & Recall 90\%                                                            & 0.764           & 1.082           & 0.044              & 5.482           & 5.066       & 0.971        \\
                                         & Recall 80\%                                                            & 0.765           & 1.083           & 0.044              & 5.522           & 5.099       & 0.971        \\
                                         & Recall 70\%                                                             & 0.768           & 1.091           & 0.045              & 5.571           & 5.106       & 0.971        \\
                                         & Recall 50\% temporal                                                   & 0.773           & 1.107           & 0.046              & 5.933           & \textbf{5.359}       & 0.971        \\
                                         & No detected agents                                                     & 0.803           & 1.151           & 0.057              & 6.113           & 5.311       & 0.971       \\
\hline
\end{tabular}
\end{table*}

The motion prediction task usually assumes perfect perception, i.e., that the ground-truth past trajectory of all actors is given and that the map information is available. However, in practice, self-driving vehicle perception systems have noise that will translate into false negatives, false positives, and incomplete agent tracking or id switches. 
We therefore conduct a robustness study, in which perception system noise is simulated.  

We perform an ablation for PGP \cite{deo2021multimodal} as having an object detector with 80\%,  90\%, and 95\% recall -- the latter being a realistic setup under good conditions. This ablation is done in an independent manner for dynamic objects detection and lane detection.

Each agent or lane segment is masked following a Bernoulli distribution with a probability of 20\%, 10\% and 5\% respectively. 
Table \ref{tab:robustness} shows the results for both lanes and dynamic agents in terms of $minADE_k$, $miss\_rate_k$ -- metrics used for ranking nuScenes benchmark -- and Behavioral Cloning (BC). \footnote{ \href{https://github.com/sancarlim/Explainable-MP/tree/v1.1}{Robustness Analysis code repository}.} 

We found PGP model to be quite robust to failures in the detection of dynamic agents. However, when we introduce noise in the perception of lanes, performance is drastically decreased, specially in the BC metric, given the importance of the lanes for this metric. 
This is probably due to the fact that this model largely exploits lane information as a strong inductive bias, which contains both the direction of traffic flow  and legal paths for each agent. 

We perform a second experiment where we analyze the effect of not detecting some frames of the agents that interact with the focal agent. Results are shown in the second last row.  We consider an interaction if their trajectory intersects with a 20 m radius of the target agent. Each frame has a 50\% probability of being detected. Results do not show a noticeable drop in performance in this case either. Finally, we perform a final experiment where we simulate that no dynamic agent is detected. Results are shown in the last row.
Given these results, we can state that the model relies heavily on lane information as well as on the focal agent's past trajectory.

In order to evaluate if these results generalize to a different scenario and model, we perform the same experiments for DenseTNT in Argoverse dataset with K=12. Table \ref{tab:robustnessDenseTNT} shows the results of the robustness analysis for DenseTNT.
In the same way as in the previous scenario, failing to perceive the lanes has a more detrimental effect that failing to perceive the dynamic agents in terms of accuracy. Again, the masking of all agents in the scene decreases the performance, but it still maintains a reasonable level.
Noise in the perception of lanes reduces diversity, while masking of dynamic agents seems to increase diversity -- see second-last row in Table \ref{tab:robustnessDenseTNT} . The reason for this behaviour is that the fewer lanes perceived, the fewer plausible trajectories. Not perceiving certain agents, however, could cause the AV to go to an occupied space, increasing the diversity in the prediction at the cost of raising the risk of a collision. Admissibility decreases when lanes are masked for the same reason, while not perceiving agents in the scene has no effect on this aspect, as expected.  

The behavior observed in this analysis can have several explanations. First, the historical information of the focal agent in terms of position, velocity and acceleration is already very informative when inferring the motion of the vehicle. 
Second, most scenarios lack interactions that drastically impact the future trajectory of the focal agent. Even though nuScenes is one of the most complex trajectory forecasting datasets and it includes several highly interactive scenarios, this is still insufficient.
This is one of the main weaknesses of current motion prediction benchmarks. In the most safety critical situation, other road users should play a crucial role. However, this is not reflected in the current benchmarks, which should cover a wide range of edge cases. These rare occurrences are easily missed and thus are often missing in datasets. Humans are naturally proficient at dealing with these extreme cases, but this is not true for autonomous systems. Therefore, we need to deal with it carefully. 
Another conclusion we can draw from this analysis is the importance of dynamic object detection. Object detectors are usually  trained on single images. However, this can be sub-optimal. Humans exhibit an understanding of the dynamics of scenes. If objects move against a static background, we can detect them quite easily, despite darkness, rain or other occlusions. A dynamic object detection system would prevent missing objects in specific frames, thus avoiding dragging or carrying this error into the next stages of the pipeline.



\section{Intention prediction}
\label{sec:intention}
\setcounter{figure}{0}  



Understanding the intention of the surrounding road agents is most relevant to mid- and long-term prediction and decision making. In order to drive through dynamically changing traffic scenarios, a multi-modal intention prediction module is necessary to adapt to the different scenarios. 
Intention prediction differs from motion or trajectory prediction in that it corresponds to discrete high-level behaviors, semantically different from each other, which we can consider modes of the future trajectory distribution.
Many trajectory prediction models are not inherently multi-modal and suffer from mode collapse. We wish to disambiguate the output and disentangle these modes into clear high-level intentions. In addition, this may contribute towards the interpretability of the overall system since it is more aligned to how humans think while driving.

 
Another potentially important advantage of intention prediction is that it is much less sensitive to the actions of the AV, compared to the more detailed task of trajectory prediction in which the precise future trajectory of the agent will depend on the actions of the AV if they are "interacting". This implies a greater potential for the intention prediction task to be dealt with in an open-loop manner. 

Most studies in the field of motion prediction work on trajectory prediction and only a few on intention prediction \cite{intention1, intentnet}, which frame the task as a classification problem. However, these rely on predefined trajectories obtained by hand-crafted principles, failing to capture comprehensive representations for the future distribution. These methods lack a consistent evaluation, which makes them suboptimal and lag behind state-of-the-art regression and generative models. In this section, we explore a new formulation for the intention prediction framework, by using a simple post-hoc approach that could be added on top of different state-of-the-art multi-modal motion prediction methods.  

 We extend the DenseTNT model \cite{Gu2021DenseTNTET} to instead perform intent prediction~\footnote{ \href{https://github.com/sancarlim/DenseTNT-Intent/tree/v1.0}{DenseTNT-Intent code repository}.}.
 We observed that the set of K potential output goals for the focal agent are not inherently different. Oftentimes, they are right next to each other or almost overlapping. The reason for this is most likely that there is uncertainty in the motion profile. It would be more convenient and intuitive if each of the K locations were different, disentangling the longitudinal uncertainty from the intention.
Another problem we detected in this model is that often the predicted goals are not admissible or rule abiding, falling off the road or in lanes going in the opposite direction. It would also be desirable to have a probability associated with each mode, to facilitate the subsequent decision making.

A simple approach to solve this would be to cluster the different outputs of DenseTNT into intentions. 
As a first step, we trained DenseTNT to output 12 goals instead of 6 -- by default -- optimizing the miss rate metric instead of FDE, since we believed these two changes would lead to better coverage of the output distribution. Indeed, it showed higher diversity and a reduction of failure modes. 

The model outputs a set of goals $\mathcal{G}=\{g_1,\dots,g_N\}$, where each goal $g_n\in\mathbb{R}^2$ is a point in birds-eye-view. In order to capture different intentions, we cluster the goals, forming a set of clusters $\mathcal{C}=\{c_1,\dots,c_K\}$. Each cluster, $c_k$, contains three components,
\begin{align}
    c_k = (\gamma_k, \mu_k, \Sigma_k)\enspace,
\end{align}
where $\gamma_k\in[0,1]$, with $\sum_k\gamma_k = 1$, is the probability of a goal ending up in cluster $k$. $\mu_k\in\mathbb{R}^2$ is the mean position of the cluster and $\Sigma_k\in\mathbb{R}^{2\times 2}$ is the covariance of the cluster. Note that we expect the covariance to be high along the lane, corresponding to the variability in speed between agents, and low orthogonally to the lane. We propose a straightforward and intuitive approach to find $\mathcal{C}$.

\begin{table*}[]
\centering
\caption{Quantitative results for DenseTNT (12 modes) and DenseTNT-intent in terms of diversity, admissibility, and accuracy. Improvements are indicated by arrows. The first two rows show results for the whole validation set. The third and fourth row consider top 3 predictions. The last two rows describe the results for the subset of scenes with more than one plausible mode.}
\label{tab:intent}
\resizebox{\textwidth}{!}{%
\begin{tabular}{|c|c|cccc|cc|cc|}
\hline
\multirow{2}{*}{\begin{tabular}[c]{@{}c@{}}Evaluation\\ Method\end{tabular}} &
  \multirow{2}{*}{Model} &
  \multicolumn{4}{c|}{Accuracy} &
  \multicolumn{2}{c|}{Diversity} &
  \multicolumn{2}{c|}{Admissibility} \\
 &
   &
  p-minADE ($\downarrow$) &
  p-minFDE ($\downarrow$) &
  p-avgFDE ($\downarrow$) &
  p-MR ($\downarrow$) &
  p-RF ($\uparrow$) &
  $\sigma_\mathrm{yaw}^2 (\uparrow$) &
  DAC ($\uparrow$) &
  OTD (\%) ($\downarrow$) \\ \hline
\multirow{2}{*}{\begin{tabular}[c]{@{}c@{}}Whole\\ k=12\end{tabular}} &
  DenseTNT &
  2.71 &
  \textbf{3.03} &
  8.03 &
  0.79 &
  \textbf{2.65} &
  \textbf{0.096} &
  0.97 &
  6.99 \\
                                                                       & DenseTNT-Intent & \textbf{1.67} & 3.07 & \textbf{4.45} & \textbf{0.60} & 1.45 & 0.034 & \textbf{0.99} & \textbf{0}          \\ \hline
\multirow{2}{*}{\begin{tabular}[c]{@{}c@{}}Whole\\ k=3\end{tabular}} & DenseTNT        & 2.25 & 3.25 & 4.67 & 0.73 & 1.43 & 0.020 & 0.021 & 6.98       \\
                                                                       & DenseTNT-Intent & 1.67 & \textbf{3.09} & 4.45 & 0.60 & \textbf{1.45} &\textbf{ 0.037} & 0.036 & 0          \\ \hline
\multirow{2}{*}{\begin{tabular}[c]{@{}c@{}}Subset\\ k=12\end{tabular}} & DenseTNT        & 2.32 & 3.39 & \textbf{4.93} & 0.73 & 1.44 & 0.020 & 0.98  & 7.08       \\
                                                                       & DenseTNT-Intent & 1.89 & 3.28 & 5.43 & 0.68 & \textbf{1.66} & \textbf{0.058} & 0.98  & 0 \\ \hline
\end{tabular}%
}
\end{table*}

\textbf{Cluster creation}: Clustering is an NP-hard problem and commonly used algorithms, such as K-means clustering or expectation maximization, are computationally expensive and in some cases prone to degenerate solutions. In our setting, however, we have a good partitioning of the bird's-eye-view plane available already: namely the lane segments in the HD-map. We therefore propose to create clusters based on the lane segments. 
In the following, we describe the heuristic used to compute the clusters.
For each predicted goal $g_n$:
    \begin{enumerate}
        \item Obtain all lane segments \begin{math}L_n=\{l_1,\dots,l_M\}  \end{math} that intersect with a radius $r=20 m$ from the agent based on the Manhattan distance. 
        The bounding boxes of small point clouds (lane centerline waypoints) are precomputed in the map. If no lanes are found, we double the search radius $r$. 
        \item Estimate the confidence $\delta$ of each lane based on the distance to its closest waypoint, $p_{nm}$. Keep those lanes whose closest waypoint lies within a threshold radius $t$ of 2.5m. 
        \begin{equation}
        \begin{split}
        \delta_{nm} &= 1 - \frac{\min(\|g_n - p_m\|_2)}{t} \\
        \end{split}
        \end{equation}   
    \item Compute the angle between the agent's heading for mode $n$ and the lane direction, $a_{nm}$. Discard those lanes whose direction differs from the agent's heading by more than 45 degrees. This ensures that we cluster only those goals that follow the current lane direction.  
\end{enumerate}

Formally, let $\alpha_{nk}$ correspond to the probability that the goal $g_n$ belongs to cluster $c_k$. We intentionally let the assignment be soft. This makes it possible for a single goal to be assigned to multiple clusters, which is sensible under uncertainty. 
Taking the retrieved lanes $L_n$ for each goal $g_n$, we compute $\alpha_{nk}$ as follows:
    \begin{enumerate}
        \item If two lanes in $L_n$ do not merge or one is the successor of the other, then they belong to different clusters.
        \item The probability $\alpha_{nk}$, where lane $l_m$ belongs to cluster $c_k$, is computed based on the distance to the closest waypoint $p_{nm}$ and the angle  $a_{nm}$. 
        \begin{equation} 
        \begin{split}
        p_{nk}&= \delta_{nm}  \frac{1}{a_{nm}} , \\
        \alpha_{nk} &= \frac{e^{p_{nk}}}{\sum_k e^{p_{nk}}} .
        \end{split}
        \end{equation}  
    \end{enumerate}
    
This heuristic is computationally efficient and guarantees that different intentions -- such as staying on lane versus cutting out of the lane -- are represented by different clusters. 
We repeat the process for each goal $g_n$, grouping the lanes that belong to the same cluster

Then, we compute
\begin{align}
    \mu_k &= \frac{\sum_n \alpha_{nk}g_n}{\sum_n \alpha_n}\enspace,\\
    \Sigma_k &= \frac{\sum_n \alpha_{nk} (g_n - \mu_k)(g_n - \mu_k)^T}{\sum_n \alpha_{nk}}\enspace.
\end{align}
In order to compute the cluster probability, $\gamma_k$, we exploit the heatmap produced by DenseTNT. Taking the score of goal $g_n$, $s_n$:
\begin{equation} 
    \begin{split}
        p_k & = \sum_n s_n  \alpha_{nk},  \\
        \gamma_k &= \frac{e^{p_k}}{\sum_k e^{p_k}} .
    \end{split}
\end{equation} 
\textbf{Cluster visualization}: For the survey study, we visualize the clusters by considering a \textit{hard} clustering, assigning each goal to its most probable cluster.

In Figure \ref{fig:intent}, we showcase different scenarios.
The AV is represented in red and the focal agent, whose trajectory we want to predict, is represented in green. Other agents are represented with blue dots. The ground truth trajectory is shown in green. 

In the first row, the outputs of the DenseTNT model are visualized in orange. The final predicted goals are represented by orange stars. 
The arrows show the direction of the lanes.

The clustered output is shown in the second row.  The probabilities of the output trajectories are mapped to the colored bar shown on the right. The averages of the intention clusters, whose probability follow the same mapping,  are represented by coloured circles. The uncertainty in the motion profile for each intention is represented by the confidence ellipse of its covariance, visualized as a shaded contour.  
As expected, the covariance is high along the lane direction and low orthogonally, showing higher uncertainty in velocity. To deal with cases where only one goal is assigned to a cluster, we apply a fixed covariance based on our previous knowledge of 2m along the lane and 0.05m orthogonal to the lane (left-most lane cluster in Figure \ref{fig:intent}).  

\setcounter{figure}{-3}  
\begin{figure*}[!h]
     \centering
     \subfigure[]{
        \includegraphics[width=0.23\linewidth]{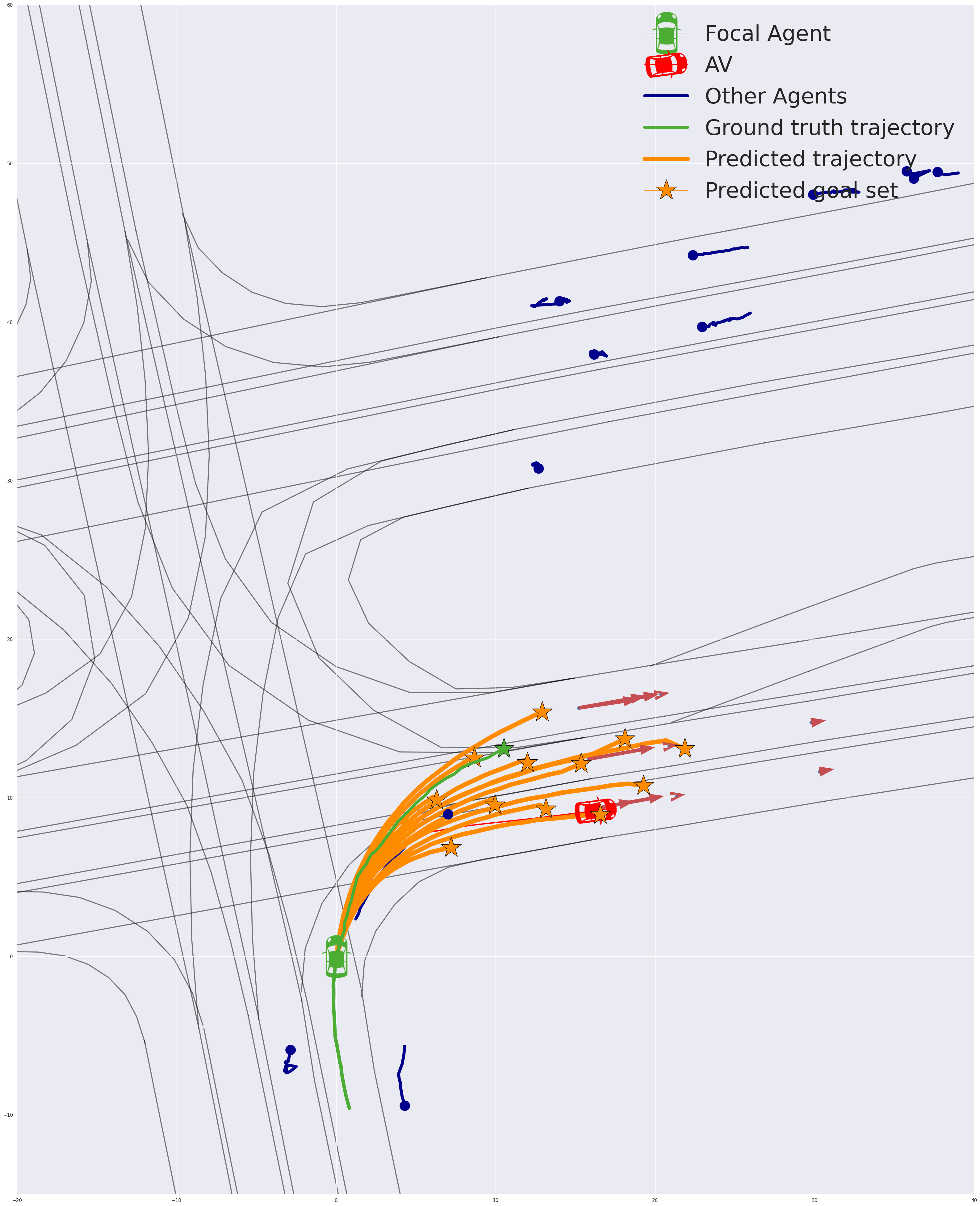}
        \phantomcaption\label{figAa}   
    }
    \subfigure[]{
        \includegraphics[width=0.23\linewidth]{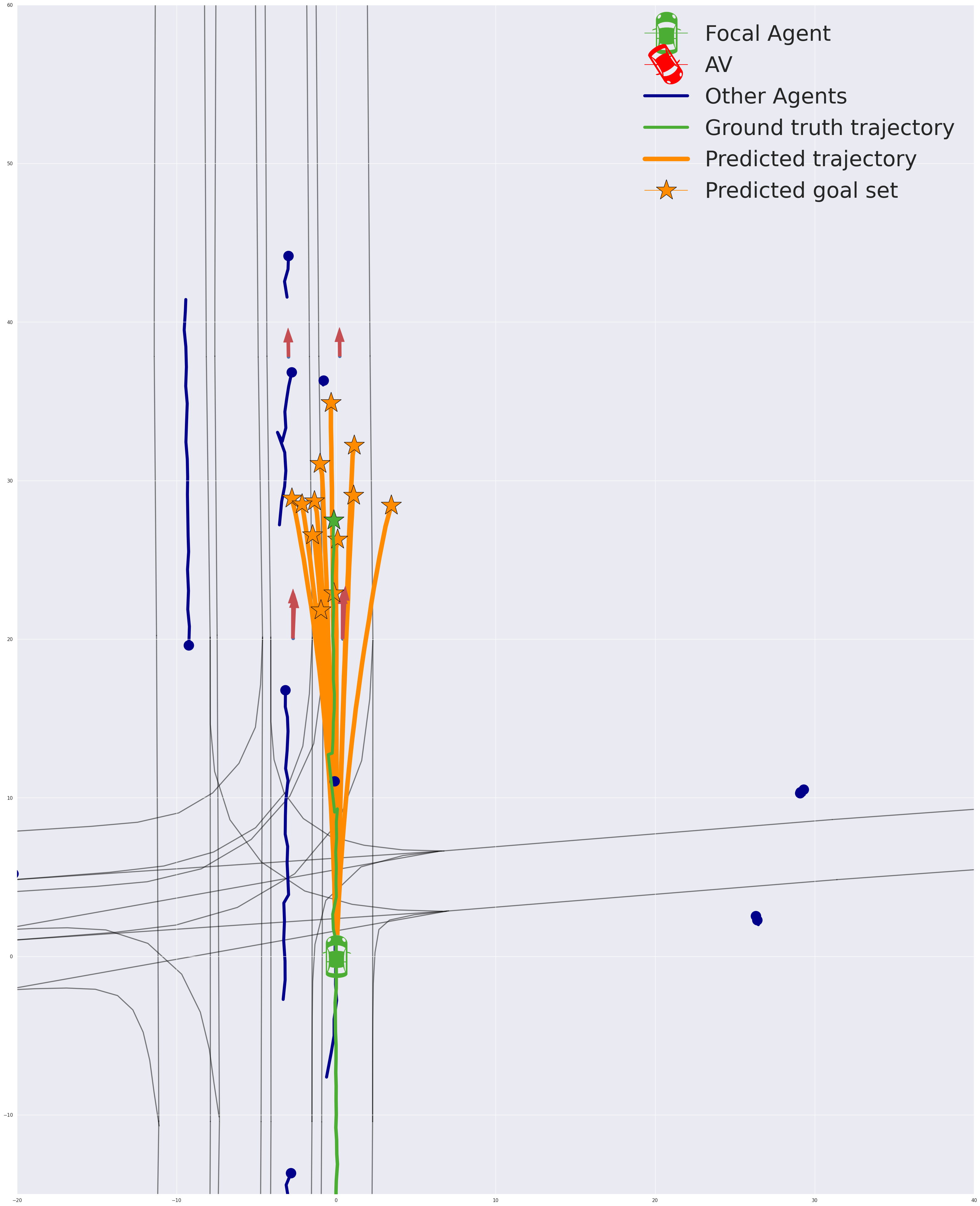} 
        \phantomcaption\label{figBa}
    }
    \subfigure[]{
        \includegraphics[width=0.23\linewidth]{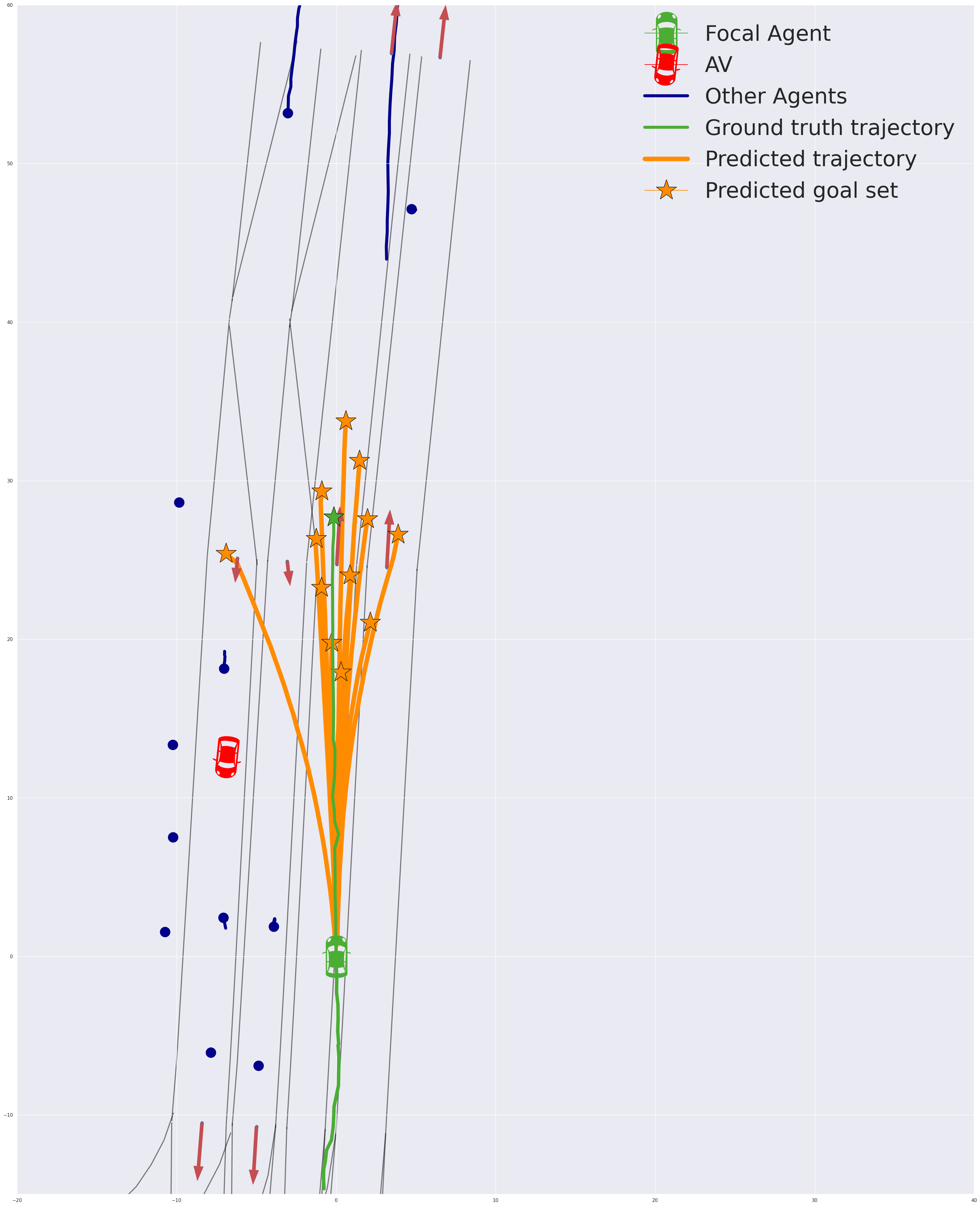}
        \phantomcaption\label{figCa}
    }
    \subfigure[]{
        \includegraphics[width=0.23\linewidth]{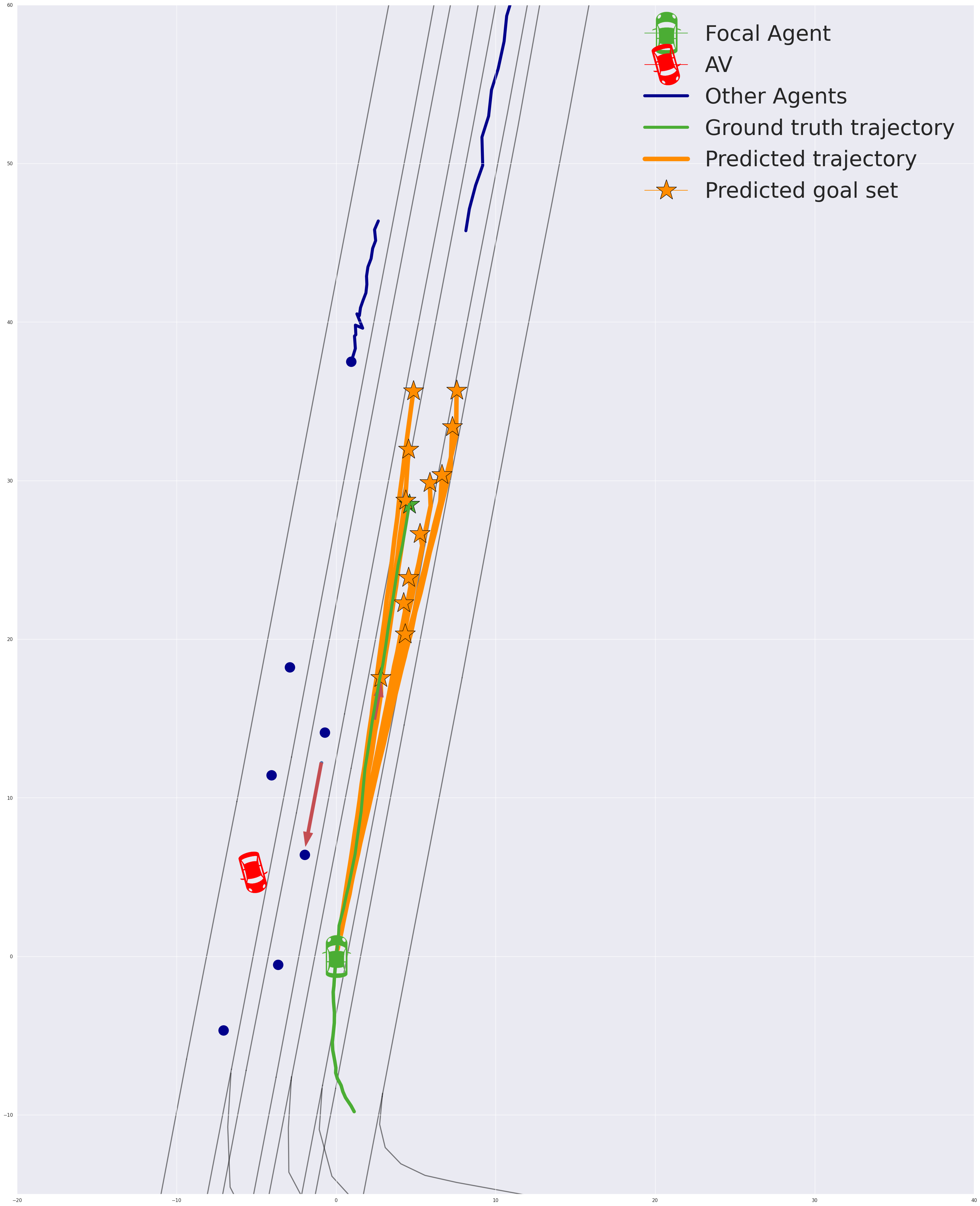} 
        \phantomcaption\label{figDa}
    } 
     \subfigure[]{
        \includegraphics[width=0.23\linewidth]{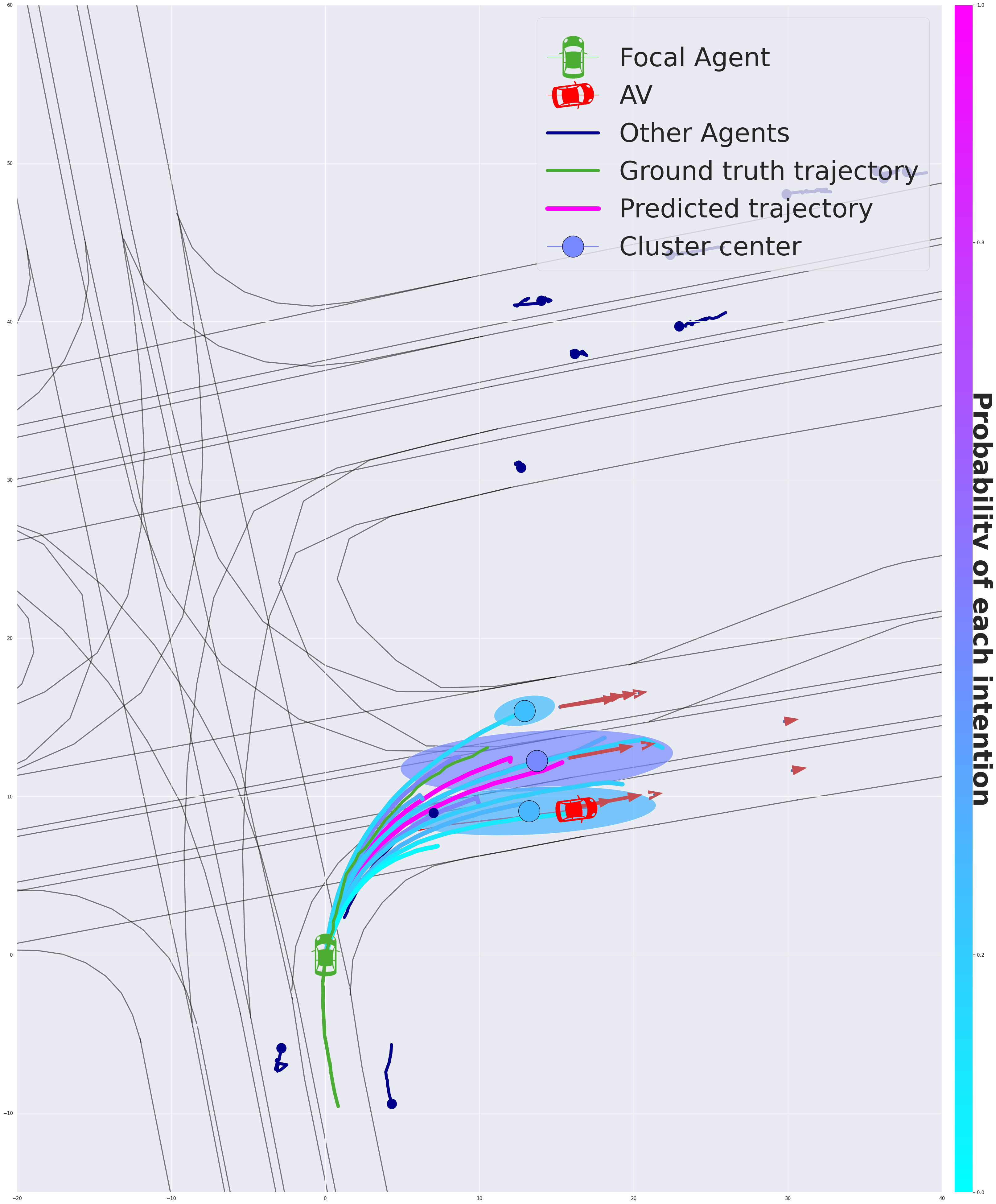}
        \label{figA}   
    }
    \subfigure[]{
        \includegraphics[width=0.23\linewidth]{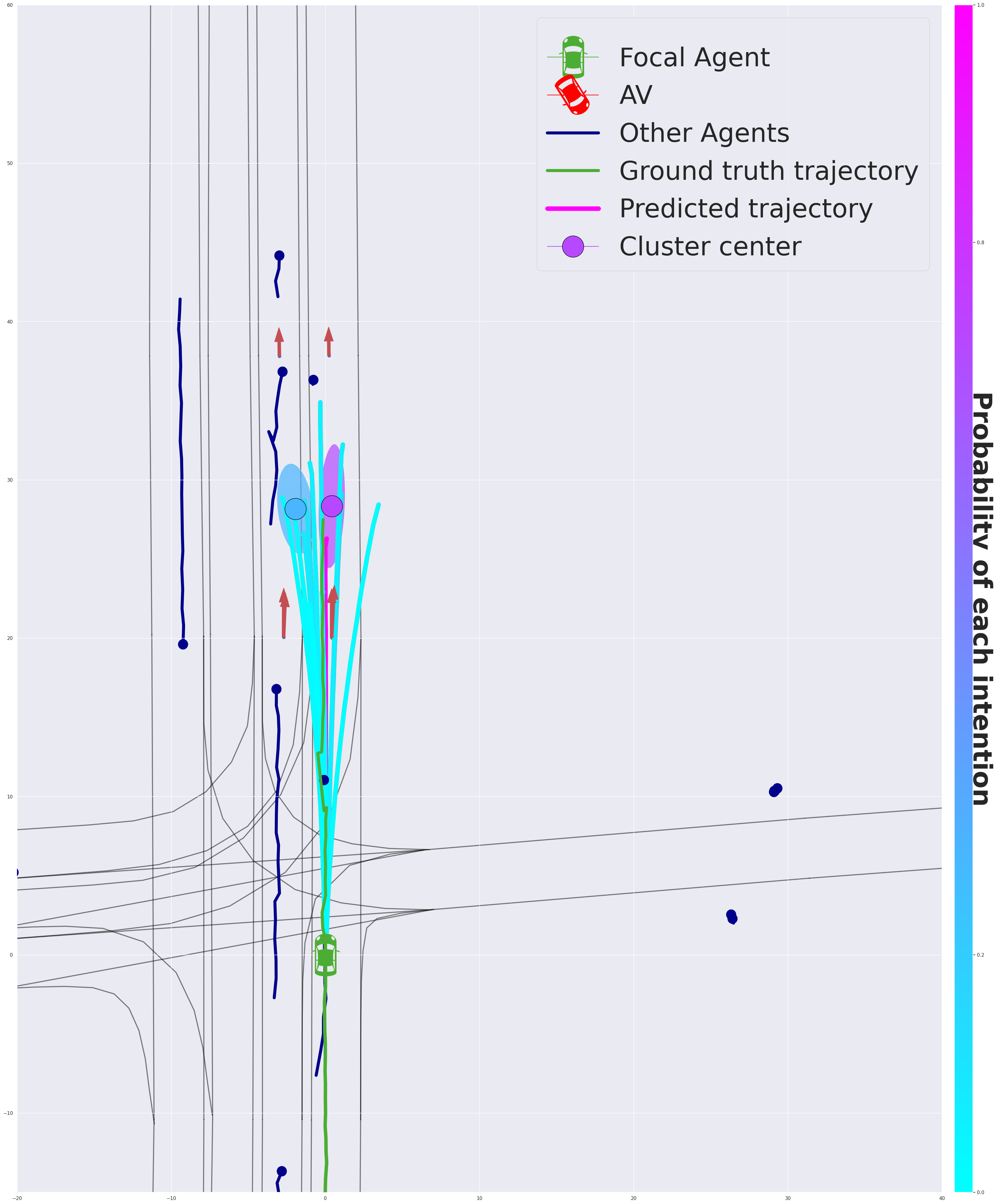} 
        \label{figB}
    }
    \subfigure[]{
        \includegraphics[width=0.23\linewidth]{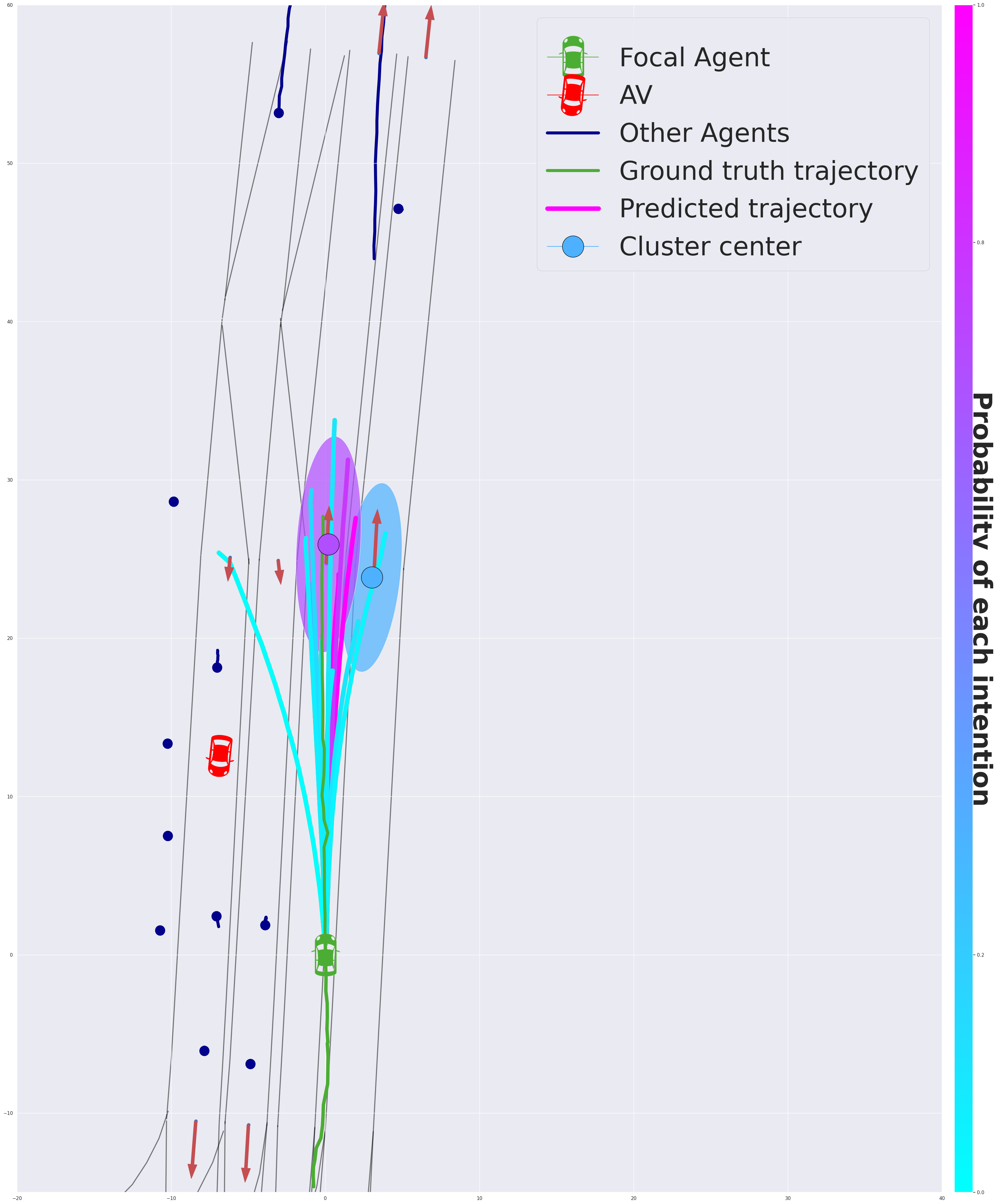}
        \label{figC}
    }
    \subfigure[]{
        \includegraphics[width=0.23\linewidth]{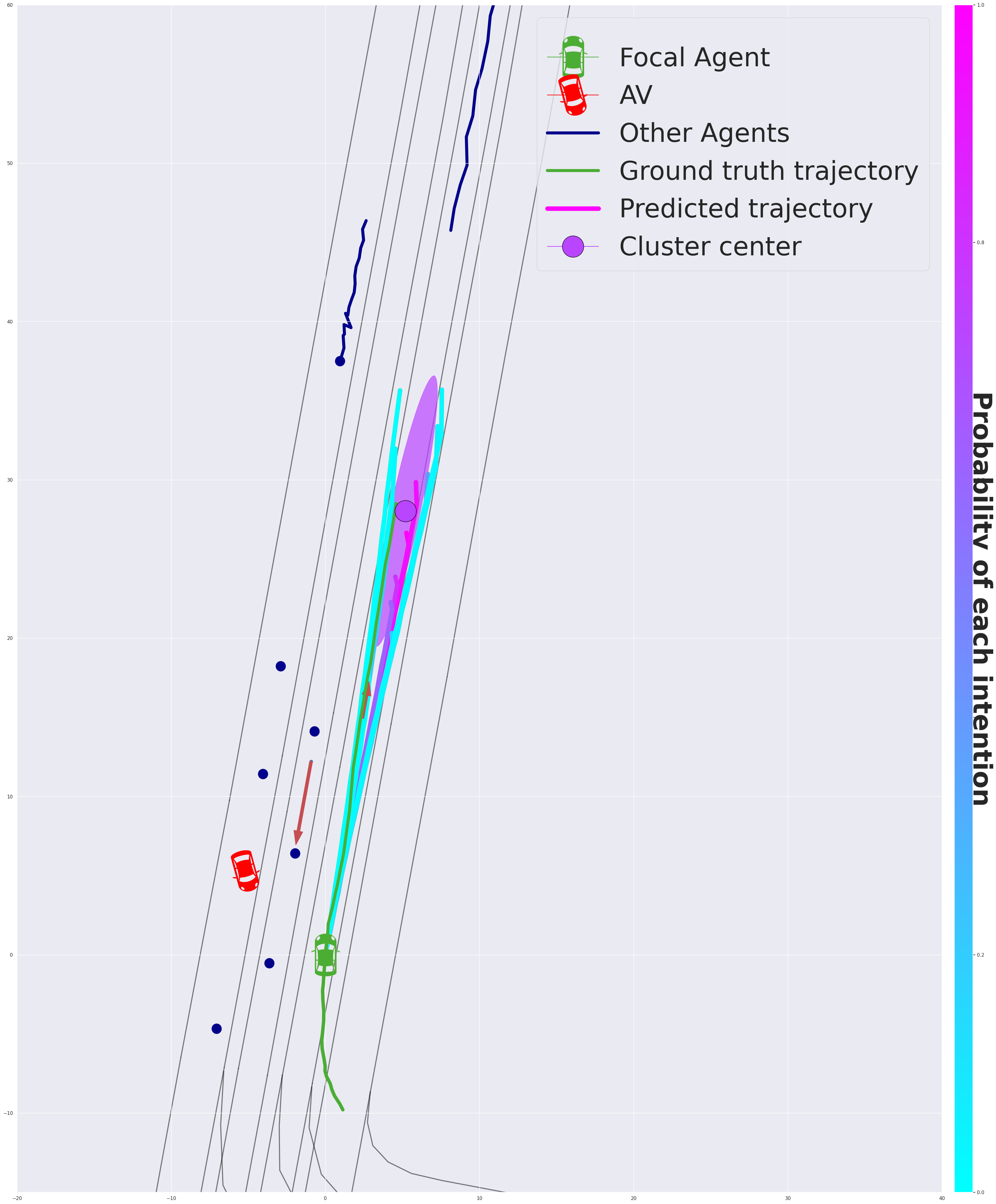} 
        \label{figD}
    } 
        \caption{\textbf{Qualitative evaluation}: DenseTNT-Intent (\textit{second row}) clusters  the output of DenseTNT (\textit{first row}) into high-level intentions with a probability associated with each cluster following the colorbar on the right. The circles represent the cluster average. Shaded contours represent the uncertainty in the motion profile. Spurious goals going off the road or in the direction of oncoming traffic are discarded.}
    \label{fig:intent}
\end{figure*}

In the first column, three intentions are detected, one for each lane, being the second the most probable one.
In the second column, the model outputs two intentions: follow and change lane. The goal that go off the road is discarded and not considered for the clustering.
In the third case, we can observe how this method eliminates spurious goals that do not follow traffic rules, going to lanes that go in the opposite direction. 
In the last scenario, the model outputs fall in both boundaries of the same lane. In this case we only output one intention.
Note that this is fundamentally different from a heatmap representation, where the output is a dense grid of probabilities that assigns the likelihood of the agent being at each position on the map at a given time. While heatmaps can provide valuable information about the distribution of possible trajectories, they do not provide a clear indication of the underlying high-level intentions of the agent. This is a fundamental limitation when it is essential to disentangle different behavioral modes and predict the agent's intended goal or destination accurately. By contrast, our proposed clustering approach allows us to identify the most probable high-level intentions of the agent, providing a more interpretable and actionable output.

\textbf{Quantitative evaluation}: This method not only disambiguates the multi-modal output, showing the possible intentions predicted by the model, but also improves the compliance of the output predictions. The clustering discards non-plausible goals that violate traffic rules, going off-road or in the direction of oncoming traffic. 

To assess the veracity of this statement, we provide measures of DAC and OTD for both DenseTNT and our method, which we call DenseTNT-Intent.  

We compute two metrics of diversity: p-RF and variance of the final heading for the different output modes,  $\sigma_\mathrm{yaw}^2$ for the whole output -- 12 modes -- and the 3 most probable outputs. We propose to modify RF and compute instead the ratio of the probabilistic versions of avgFDE and minFDE, p-RF. p-avgFDE is computed following the same heuristic as p-minFDE. For each sample with K trajectories:  

\begin{equation}
    \text{p-avgFDE} = \frac{1}{K} \sum_k FDE_k + \min(-\log p_k, -\log 0.05)
\end{equation}

Finally, we also provide probabilistic precision metrics to compare both models in terms of accuracy. All metrics must be considered simultaneously in order to get a holistic interpretation. The results are illustrated in Table \ref{tab:intent}.  

DenseTNT diversity metrics are better when evaluated with the whole output, i.e., 12 predictions. This is probably due to the fact that clustering removes bad predictions, which makes the variance of the yaw lower. Following the same intuition, avgFDE is higher for DenseTNT whereas minFDE is lower. Therefore, the ratio RF is also higher for DenseTNT. However, RF only measures diversity  without looking at the quality of this output.  Moreover, 90.7\% of the target agent's maneuvers go in a straight line. In most of these cases, all the uncertainty is in the motion profile, not in the intention, thus having only one output cluster. Figure \ref{figD} shows a clear example where the variance will be much higher for DenseTNT before the clustering. 

We decide to evaluate this hypothesis with two more evaluation scenarios.
First, we evaluate the model in the case where we only keep the three most probable predictions (k=3). In this scenario, the clustered output shows higher diversity both in terms of  $\sigma_\mathrm{yaw}^2$ and  p-RF. 
Secondly, we evaluate it in a subset of the validation set, considering only those scenes whose underlying distribution has more than one plausible mode. As expected, we encounter best results in terms of diversity in this evaluation scenario. 
For the analysis of accuracy and admissibility results, we will focus on the whole dataset wit K=12 since there are no major changes in the other dimensions.

The admissibility metrics show that clustering the output trajectories into high-level behaviors provides a more compliant prediction, with no clusters going in the direction of oncoming traffic and a 2pp  higher  DAC. We find that almost 7\% of the predictions go in the opposite direction.

In order to get a holistic evaluation, we also explore how the clustering affects the results in terms of accuracy. When evaluating the probabilistic versions of distance-based metrics, DenseTNT-Intent seems to be superior in terms of p-minADE, p-avgFDE and p-MR. This is due to the fact that it provides a lower amount of modes with a higher probability for the ground-truth intention. However, since   DenseTNT outputs more trajectories with higher variability in the motion profile, it is more likely that one of these trajectories will be closer to the ground-truth prior to the clustering, thus having a lower p-minFDE for DenseTNT. 

These results suggest that DenseTNT-Intent achieves a more scene-compliant output, in the form of intention prediction, covering the modes of the output distribution while improving the overall quality of the predictions. 
This output can be used to improve the safety and reliability of autonomous vehicles by enabling better decision-making and more accurate prediction of the agents' future behavior.

\section{Interpreting multi-modal motion prediction outputs}
\label{sec:survey}



In this section, we test our hypothesis that appropriate visualization of multi-modal trajectory representation increases the interpretability of the prediction system. We verified our null hypothesis by creating a survey in which we explore how the visualizations of multi-modal predictions, with their associated probabilities and different number of modes and prediction horizons, impacts the interpretability of the prediction system output, and how this, consequently,  might impact transparency and user trust in the overall system.

 \subsection{Methodology}
 We recruited 39 technical experts to participate in a twenty-minute survey designed within-subjects. Twenty-three percent of the subjects are women, which is similar to the inherent bias in the technical field. 
 In this survey, technical interpretability was assessed using different visualization experiments of the information provided by the prediction systems.  

The survey is divided into two sections. The visualizations from the first section are generated with the predictions of PGP on the nuScenes Trajectory Prediction dataset. The second section is generated with predictions of DenseTNT on Argoverse Motion Forecasting dataset. We choose the two most widely used datasets in the field of multi-modal motion prediction and the two models that ranked at the top of their benchmarks. This was done to increase the diversity of the study. In addition, it introduces differentiation in terms of the types of multi-modal traffic prediction frameworks, as DenseTNT utilizes target set prediction and PGP employs lane graph sampling. 

Each experiment in the survey is explored with three types of questions. First, an A/B test, in which the subject must choose between two presented cases. Second, a Likert scale, in which  a five-point scale is used to allow the individual to express how much they agree or disagree with a particular statement. Finally, an open-ended question where they can explain the reason for their answers. This final question helps us to see things from the respondent's perspective, as we get feedback in their own words. 
Please, refer to the supplementary materials for more details~\footnote{ \href{https://forms.gle/yqGjwi8cH1KKA7KM7}{Link to the survey}}. 
In the following schema, we describe the structure of the survey and the variables under study:
\begin{enumerate}
    \item Section 1. PGP visualizations in nuScenes dataset.
    \begin{enumerate}
        \item Introduction and baseline visualization with no predictions.
        \item Case 1. Single mode vs muli-mode.
        \item Case 2. Number of modes: 10 modes vs 3 modes.
        \item Case 3. Prediction time horizon: 6s vs 2s.
        \item Case 4. Importance of cameras view as an additional visualization.
    \end{enumerate} 
    \item Section 2. DenseTNT and DenseTNT-intent visualizations in Argoverse dataset.
    \begin{enumerate}
        \item Introduction and baseline visualization with no predictions.
        \item Case 5. Single mode vs multi-mode.
        \item Case 6. Trajectory prediction vs intention prediction.
    \end{enumerate} 
\end{enumerate}

These cases are designed taking into consideration the factors explored in this work, as well as those aspects that are not consistent among different evaluation benchmarks.

 \begin{table*}[]
\centering
\caption{Evaluation of survey results: Descriptive and inferential statistics for each of the cases under study. k is the number of modes, being n>1 the multi-modal scenario. We provide mean, variance, and mode as descriptive statistics. For inferential statistics, test statistic, p value and degrees of freedom (dg) based on Welch's test are shown. nS refers to nuScenes, AV refers to Argoverse dataset.}
\label{tab:survey}
\begin{tabular}{c|cc|cc|cc|cc|cc|}
\cline{2-11}
 &
  \multicolumn{2}{c|}{k=1 vs k>1 (nS)} &
  \multicolumn{2}{c|}{k=3 vs k=10} &
  \multicolumn{2}{c|}{6s vs 2s} &
  \multicolumn{2}{c|}{k=1 vs k>1 (AV)} &
  \multicolumn{2}{c|}{Trajectories vs Intentions} \\ \cline{2-11} 
 &
  \multicolumn{1}{c|}{k=1} &
  \textbf{n>1} &
  \multicolumn{1}{c|}{k=3} &
  k=10 &
  \multicolumn{1}{c|}{\textbf{6s}} &
  2s &
  \multicolumn{1}{c|}{k=1} &
  n>1 &
  \multicolumn{1}{c|}{Trajectories} &
  \textbf{Intentions} \\ \hline
\multicolumn{1}{|c|}{Mean} &
  \multicolumn{1}{c|}{3.410} &
  4.179 &
  \multicolumn{1}{c|}{3.821} &
  3.641 &
  \multicolumn{1}{c|}{4.333} &
  2.872 &
  \multicolumn{1}{c|}{3.282} &
  3.589 &
  \multicolumn{1}{c|}{3.205} &
  4.102 \\
\multicolumn{1}{|c|}{Var} &
  \multicolumn{1}{c|}{0.879} &
  0.458 &
  \multicolumn{1}{c|}{0.835} &
  1.025 &
  \multicolumn{1}{c|}{0.3333} &
  0.6932 &
  \multicolumn{1}{c|}{0.997} &
  1.143 &
  \multicolumn{1}{c|}{0.693} &
  0.568 \\
\multicolumn{1}{|c|}{Mode} &
  \multicolumn{1}{c|}{3} &
  4 &
  \multicolumn{1}{c|}{4} &
  4 &
  \multicolumn{1}{c|}{4} &
  3 &
  \multicolumn{1}{c|}{3} &
  4 &
  \multicolumn{1}{c|}{3} &
  4 \\ \hline
\multicolumn{1}{|c|}{t-Stat} &
  \multicolumn{2}{c|}{-4.139} &
  \multicolumn{2}{c|}{0.822} &
  \multicolumn{2}{c|}{9.007} &
  \multicolumn{2}{c|}{-1.313} &
  \multicolumn{2}{c|}{-4.989} \\
\multicolumn{1}{|c|}{p-value} &
  \multicolumn{2}{c|}{$9\cdot10^{-5}$} &
  \multicolumn{2}{c|}{0.41} &
  \multicolumn{2}{c|}{$10^{-10}$} &
  \multicolumn{2}{c|}{0.193} &
  \multicolumn{2}{c|}{$1.9\cdot10^{-6}$} \\
\multicolumn{1}{|c|}{df} &
  \multicolumn{2}{c|}{69} &
  \multicolumn{2}{c|}{75} &
  \multicolumn{2}{c|}{67} &
  \multicolumn{2}{c|}{75} &
  \multicolumn{2}{c|}{75} \\ \hline
\end{tabular}
\end{table*}

\begin{figure*}
    \centering
    \includegraphics[width=1\linewidth]{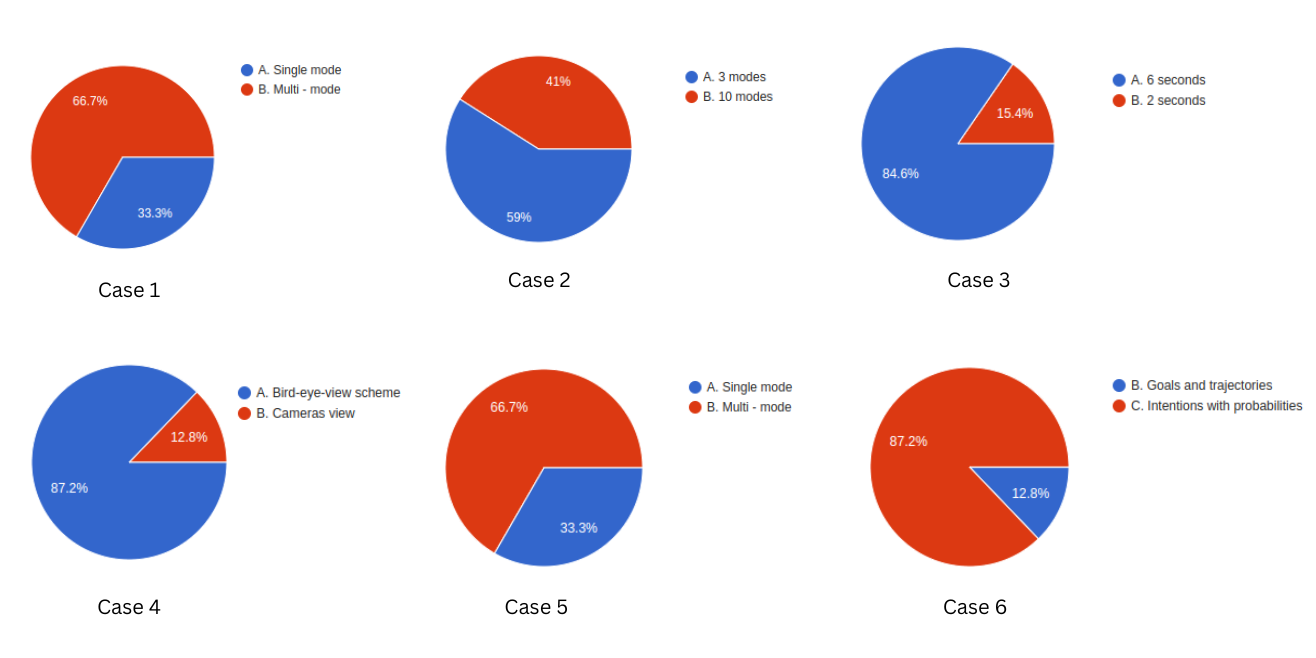}
    \caption{A/B questions pie charts}
    \label{fig:survey}
\end{figure*}

\subsection{Results} 
Table \ref{tab:survey} details the descriptive and inferential statistics. The 5-point Likert scale questions results are  analyzed as interval data, being described in terms of means, variances, and mode. Parametric analysis of ordinary averages of Likert scale data is justifiable by the Central Limit Theorem.  We choose Welch's Test to determine the validity of our hypothesis and determine the statistical significance of the difference in means for both distributions in each of the cases under study. 
Test statistic, p-value, and degrees of freedom are described in table~\ref{tab:survey}.
The pie charts for the A/B type questions are shown in figure~\ref{fig:survey}.
In the following, we detail our findings in each of the surveyed scenarios.

\textbf{1. Single mode vs multi-mode predictions.}
In this first scenario, evaluated with PGP model on nuScenes dataset, we test our hypothesis that multi-modality increases the interpretability of the prediction system output over single-mode predictors.
20\% of participants preferred single mode over multi-mode prediction representation in terms of interpretability of the predictive system. 
These subjects report finding the multi-modal visualizations more confusing, yet more reliable for the system.

We found a statistically significant difference of 0.8 points in the interpretability scoring for multi-modal predictions versus single-mode predictions, with a p-value of $9\cdot10^{-5}$. This is consistent with our hypothesis that multi-modality increases interpretability of the output of prediction systems. 

\textbf{2. Three modes versus ten modes.}
In this scenario, we study the optimal number of modes for better understanding of the predictive system.
41\% of participants chose 10 modes mode over 3 modes in terms of interpretability.  
The most frequent argument is that the interpretation of ten predictions may be complex for humans, but seems more reliable for the system, increasing the confidence in it. 
When asking to rate both cases in terms of interpretability in a 5-point Likert scale, we found no significant difference in the means of both distributions.   

\textbf{3. Prediction time horizon: 6s versus 2s.}
In this scenario, we study the importance of long-term prediction over short-term prediction for the interpretability of the output.
84.6\% of the participants preferred six seconds of prediction over two seconds in terms of interpretability of the output predictions. 
Most argue that 6s provides more information in order to understand the scenario.  

In the Likert scale rating, we found a statistically significant difference of 1.5 points in the means of both distributions, with a p-value of $10^{-10}$. This is consistent with our hypothesis that longer predictions horizons increase the interpretability of the predictions output and the reliability in the system.

\textbf{4. Cameras view as additional visualization.}
In this scenario, we explored the influence of adding the visualization of six cameras placed in the AV on the interpretability of the traffic scene. 70\% of participants believe the additional visualization helps them understand the scene. 87.2\% believe that it is easier to understand the behavior of the focal agent and its predicted trajectory.

\textbf{5. Single mode vs multi-mode predictions with DenseTNT on Argoverse dataset.}
We repeat the first scenario on a different setup to evaluate the difference in scene representation of the Argoverse and nuScenes datasets, as well as the output representation of DenseTNT versus PGP model.
We found different results from those of the first scenario. In this case, 33.3\% of participants chose single mode predictions over multi-mode predictions in terms of interpretability.   
The biggest difference, however, is in the Likert scale rating. In this case, both means are almost identical, with no statistical difference between distributions.  
This notable difference is mainly due to the output representation in the case of DenseTNT, which outputs 12 trajectories with no probability associated. People believe that this makes it difficult to interpret the output. This result is yet another justification for the need of a posterior clustering into intentions with associated probabilities, which leads us to the last scenario.

\textbf{6. Trajectory prediction versus intention prediction.}
We test the hypothesis presented in Section \ref{sec:intention}. We posit that intention prediction can contribute towards the interpretability of the system since it disambiguates the multi-modal prediction output and produces a clustered output that is more in line with the way human drivers think.  
87.2\% of participants believe the clustered output with its associated probability is more understandable when visualizing the future predicted behavior of the focal agent.  

Most argument argue that they find it more useful to know the high-level behavior than the concrete future trajectory, as it is more in line with the way humans reason. Others pose it is faster to interpret and adds more information while removing the ambiguity of multi-modal trajectory prediction. 
In addition, they also indicate the importance of having uncertainties associated with the predictions, to allow the subsequent algorithm to adopt a probabilistic framework.
However, 5 participants believe that the second option is more confusing and difficult to understand.

When looking at the inferential statistics of the Likert scale question, there is a statistically significant difference of 0.9 points in the mean of both distributions, with a p-value of $3\cdot10^{-6}$. This verifies our hypothesis that intention prediction helps improve the understanding of the predictive system. 

\subsection{Discussion}

This study comes with several limitations that should be considered.
First, Likert scales are subject to distortion due to \textit{central tendency bias} (i.e., avoidance of using extreme categories), and \textit{acquiescence bias} (i.e., agreeing with statements in the survey). 

Second, the A/B type questions leave no room for a neutral response when no option is preferred. However, this could be specified in the Likert scale and in the open-ended question of the particular scenario. It has been taken into consideration for the analysis.

Finally, some subjects claimed to be unsure of the purpose of some questions and believe that some answers are highly dependent on the particular task. For example, for a developer it may be more interesting to have more information at the cost of a more complex representation. Some questions also depend on the traffic scenario, speed, and complexity of the road, with the optimal number of modes or prediction horizon being different for each case. 

Despite these limitations, we can derive some conclusions from this study.
First, it verifies our hypothesis that multi-modality is not only important for better decision making and planning a safer route, but it also improves the understanding of prediction systems when properly visualized. Another important insight is that the form of visualization is crucial and it is important to show the different probabilities associated with each trajectory in a clear way.
Second, longer prediction time horizons provide a better understanding of the whole traffic scene and make the overall system more interpretable.
Finally, the need for intention prediction is supported by the survey findings. Clustered outputs provide a more human-like representation of the predictions into high-level behaviours. 


\section{Conclusions}
\label{sec:discussion} 
In this work, we move towards the design of reliable motion prediction models based on evaluation, robustness, and interpretability of the outputs. 

Our findings can be summarized as follows:

\begin{itemize}
    \item We highlighted the main gaps and differences in current evaluation methodologies, especially in terms of lack of diversity assessment and admissibility with the traffic scene. We identify the main aspects that are critical for the evaluation of multi-modal motion prediction and propose a more comprehensive and holistic evaluation framework. 
    \item In the robustness analysis, we showed how failure to perceive the road topology has a greater impact on system performance compared to failure to perceive other agents on the road, due to the significant inductive biases introduced by the lanes. This also showcases the need for more comprehensive datasets covering complex scenes with high interaction and wide range of edge cases. 
    \item We provided DenseTNT-intent outputs with high-level intentions that prove to be diverse, compliant, and accurate, improving the overall quality of the predictions.
    \item The results of the study suggest that this new representation improves the interpretability of the output prediction. Our first hypothesis that multi-modality improves interpretability over single-mode predictions is also verified. Finally, long-term predictions appear to provide a better understanding of the predicted traffic scene. 
\end{itemize} 


The proposed approach and findings make a significant contribution to the development of trustworthy motion prediction systems for autonomous vehicles. By comprehensively analyzing current evaluation metrics, identifying gaps, and proposing a new holistic evaluation framework, this work provides a valuable foundation for future research in the field. Additionally, the formulation of a method for assessing spatial and temporal robustness, as well as the proposed intent prediction layer, demonstrate innovative solutions for addressing the complex challenges of multi-modal motion prediction. Finally, the assessment of interpretability through a survey of different visualization techniques offers further insights for enhancing the performance and transparency of autonomous vehicle systems. Overall, this work represents a substantial step forward in the design of trustworthy artificial intelligence for safe and efficient autonomous driving.

\section*{Acknowledgment}
This work was supported by HUMAINT project at the Algorithmic Transparency Unit at the Directorate-General Joint Research Centre (JRC) of the European Commission, and in part by the Spanish Ministry of Science and Innovation under Grant PID2020-114924RB-I00, and by the Community Region of Madrid under Grant S2018/EMT-4362 SEGVAUTO 4.0-CM. In addition, part of this work has been carried out during the \textit{Eye for AI Program} thanks to the support of Zenseact and AI Sweden. 

\section*{Disclaimer}
The information and views expressed in this paper are purely those of the authors and do not necessarily reflect an official position of the European Commission.

\ifCLASSOPTIONcaptionsoff
  \newpage
\fi



%


\bibliographystyle{unsrt}
\bibliography{egbib}

%

\end{document}